\documentclass[10pt,twocolumn,letterpaper]{article}

\usepackage[pagenumbers]{wacv} 

\usepackage{graphicx}
\usepackage{amsmath}
\usepackage{amssymb}
\usepackage{booktabs}

\usepackage{mathtools}

\usepackage{wrapfig}
\usepackage{caption}
\usepackage{subcaption}
\usepackage{multirow}
\usepackage{tabularx}
\usepackage{floatrow}
\newfloatcommand{capbtabbox}{table}[][\FBwidth]

\usepackage{algorithm}
\usepackage{algorithmic}
\usepackage{pifont}
\usepackage{balance}

\DeclarePairedDelimiterX{\infdivx}[2]{[}{]}{%
  #1\;\delimsize\|\;#2%
}

\usepackage[pagebackref,breaklinks,colorlinks]{hyperref}
\usepackage[accsupp]{axessibility}  

\usepackage[capitalize]{cleveref}
\crefname{section}{Sec.}{Secs.}
\Crefname{section}{Section}{Sections}
\Crefname{table}{Table}{Tables}
\crefname{table}{Tab.}{Tabs.}

\def\wacvPaperID{45} 
\def\confName{WACV}
\def\confYear{2024}

\begin{document}

\title{Latent Space Energy-based Model for Fine-grained Open Set Recognition}

\author{Wentao Bao\\
Michigan State University\\
{\tt\small baowenta@msu.edu}
\and
Qi Yu\\
Rochester Institute of Technology\\
{\tt\small qi.yu@rit.edu}
\and
Yu Kong\\
Michigan State University\\
{\tt\small yukong@msu.edu}
}

\maketitle

\begin{abstract}
   Fine-grained open-set recognition (FineOSR) aims to recognize images belonging to classes with subtle appearance differences while rejecting images of unknown classes. A recent trend in OSR shows the benefit of generative models to discriminative unknown detection. As a type of generative model, energy-based models (EBM) are the potential for hybrid modeling of generative and discriminative tasks. However, most existing EBMs suffer from density estimation in high-dimensional space, which is critical to recognizing images from fine-grained classes. In this paper, we explore the low-dimensional latent space with energy-based prior distribution for OSR in a fine-grained visual world. Specifically, based on the latent space EBM, we propose an attribute-aware information bottleneck (AIB), a residual attribute feature aggregation (RAFA) module, and an uncertainty-based virtual outlier synthesis (UVOS) module to improve the expressivity, granularity, and density of the samples in fine-grained classes, respectively. Our method is flexible to take advantage of recent vision transformers for powerful visual classification and generation. The method is validated on both fine-grained and general visual classification datasets while preserving the capability of generating photo-realistic fake images with high resolution.
\end{abstract}

\section{Introduction}
\label{sec:intro}

Open-set recognition (OSR) aims to recognize samples of known classes and reject the unknown in an open-world~\cite{ScheirerTPAMI2012,geng2020recent,BendaleCVPR2016}. Because of its importance to trustworthy AI applications such as autonomous driving~\cite{WongCoRL2020,BlumICCV2019} and medical diagnosis~\cite{PrabhuNIPS2019}, this topic has received increasing attention in recent years. However, most existing OSR literature~\cite{ScheirerTPAMI2012,ScheirerTPAMI2014,BendaleCVPR2016,yoshihashiCVPR2019,kong2021opengan,saito2021NIPS,chenPAMI2021,BaoICCV2021,vazeICLR2022} ignore the \textit{class granularity} of samples in an open-world. 
For example, identifying a \textit{bird} image as unknown by a model trained on multiple \textit{bird species} is much more difficult than identifying an \textit{elephant} image as unknown by a model trained on \textit{cat} and \textit{dog} images. 
Thus, to perform well on fine-grained recognition, class granularity needs to be valued in OSR. 

Existing OSR approaches primarily focus on open-set classes of coarse granularity, and assume that more unseen classes indicate higher openness in testing~\cite{ScheirerTPAMI2012}. 
Recently, \cite{vazeICLR2022} re-defined the openness over the inter-class similarity. 
and advocates that OSR models should perform well when the extremely hard unknowns (\ie unknown examples are the most similar to known class samples) exist in testing. Such a definition of openness implies that fine class granularity poses high openness in an open world. However, to the best of our knowledge, it has not been studied how to take the class granularity into account in OSR modeling. 

\begin{figure}
    \centering
    \includegraphics[width=\linewidth]{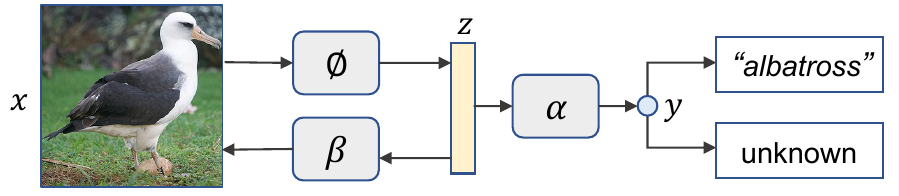}
    \caption{Hybrid modeling of discrimination and generation for open-set recognition. The $\phi$, $\beta$, and $\alpha$ are deep neural networks.}
    \label{fig:framework}
\end{figure}

The challenges of fine-grained OSR (FineOSR) lie in that the visual appearance of images from closed- and open-set classes are very similar such that the low-level details of images and class attributes are vital in model design. However, on one hand, existing fine-grained visual recognition (FGVR) methods~\cite{lin2015bilinear,DubeyECCV2018,yu2018hierarchical,zhang2021multi} are not applicable to the open set though effective on a closed set. On the other hand, to manage the separability between the known and the unknown, state-of-the-art OSR approaches~\cite{GeBMVC2017,PereraCVPR2020,kong2021opengan,chenPAMI2021} rely on deep generative models to generate the unknown, which requires high dimensionality to represent the fine-grained classes. Therefore, how to achieve good separability between fine-grained classes with awareness of class granularity in a low-dimensional space is technically challenging and under-explored in literature. 

To tackle these challenges, we resort to hybrid modeling of classification and generation for images from fine-grained classes in an open world, as shown in Fig.~\ref{fig:framework}. The intuition behind the auxiliary generation task is that the generative modeling naturally supplements the discriminative OSR objective with fine-grained features of low-level image details. Thanks to the recent advances in the energy-based model (EBM)~\cite{JEM_ICLR2020,VERA_ICLR2021}, it is promising to formulate the distinctly different objectives of discriminative and generative problems in a principled way. Moreover, to mitigate the difficulty of density estimation in a high dimensional space of EBMs, recent latent space EBMs~\cite{pangNIPS2020,pangICML2021,zhangNIPS2021,xiaoICLR2021,nijkampICLR2022} introduces the energy prior in a low-dimensional space, which pushes this line of research towards real-world practice.

Therefore, we propose to formulate the FineOSR task by a low-dimensional latent space EBM. In a nutshell, we develop the latent space EBM framework, which takes into account the \emph{expressivity}, \emph{granularity}, and \emph{density} of the fine-grained classes in an open-world. To improve the expressivity of EBM latent features, we propose an attribute-aware information bottleneck (AIB) objective to train the latent space EBM. To address the fine granularity of class representation, we explicitly disentangle the holistic latent feature into attribute-specific features by residual attribute feature aggregation (RAFA). Lastly, benefiting from the generative density estimation of EBM, we introduce the uncertainty-aware virtual outlier synthesis (UVOS) to generate virtual unknown samples in the latent space, which could reduce the open space risk of each known class. The proposed method takes advantage of the principled Bayesian probabilistic modeling such that both the fine-grained OSR and photo-realistic visual generation are simultaneously learned in an end-to-end manner.

Experimental results on fine-grained visual datasets \textit{CUB}~\cite{CUB200} and \textit{DeepFashion}~\cite{DeepFashion}, as well as the coarse-grained visual dataset TinyImageNet~\cite{le2015tiny} dataset, show that our method is competitive to existing OSR methods while preserving the capability of generating photo-realistic images. Our contributions are summarized as three-fold:
\begin{itemize}
\item We propose a novel latent space EBM for fine-grained open-set recognition (FineOSR), which aims to recognize the known and reject the unknown samples from fine-grained categories in an open world.
\item We address the expressivity, granularity, and density of the latent variables by leveraging attributes and generative density estimation for the fine-grained OSR.
\item Our experimental results show superior recognition and generation performance to existing EBMs and are competitive to the discriminative OSR baselines.
\end{itemize}

\section{Related Work}
\label{sec:related_work}

\paragraph{Open-set Recognition} The goal of open-set recognition (OSR) is to enable a classification model to be aware of the samples from unknown classes~\cite{ScheirerTPAMI2012,geng2020recent}. The early pioneering work~\cite{ScheirerTPAMI2012} proposed an ``one-versus-rest" method based on the support vector machine (SVM), which inspired a line of following-up research~\cite{ScheirerTPAMI2014,JainECCV2014,junior2016specialized}. Recent deep learning OSR methods~\cite{BendaleCVPR2016,perera2019deep,mundt2019open,BaoICCV2021,wangICCV2021} aim to learn a well-calibrated confidence or uncertainty measurement to identify the unknown. Another line of OSR approaches~\cite{GeBMVC2017,kong2021opengan,chenPAMI2021,fangICML2021,zhou2021learning,Yue_2021_CVPR,LuAAAI2022pmal,vazeICLR2022} achieve remarkable performances by leveraging generative adversarial networks (GANs)~\cite{GAN} or other generative models to synthesize the unknown in training. However, how to formulate the distinct objectives of the auxiliary generative and the discriminative tasks for the OSR problem in a principled way is not fully explored in the existing literature.

\vspace{-2mm}
\paragraph{Fine-grained Visual Recognition} The FGVR task aims to distinguish between classes by identifying subtle inter-class differences. Existing FGVR literature can be categorized into methods by localization or segmentation, end-to-end learning, and using external data~\cite{FGVR_Survey2021}. Localization-based methods leverages object detection~\cite{lin2015deep,zhang2016spda}, deep filters~\cite{zhang2016picking}, and attention mechanism~\cite{fu2017look,ji2020attention} to localize the attribute relevant features. End-to-end approaches learn the bilinear or deep CNN features and the FGVR objective through a single loss function~\cite{Lin_2015_ICCV,dubey2018maximum,zheng2019learning}. Some recent methods also exploit the external data such as web images~\cite{xu2016webly}, multi-modality data~\cite{he2017fine,zhang2018audio}, and human knowledge~\cite{chen2018knowledge}. However, only a few recent works~\cite{Dai2021,gillert2021towards} tackle the FGVR in an open-set scenario. In this paper, we follow the end-to-end FGVR paradigm under the open-set scenario such that the \emph{openness} of class granularity can be handled in a stand-alone model.

\vspace{-2mm}
\paragraph{Deep Generative Models} Generative models such as the GANs~\cite{GAN} have achieved remarkable progress in recent years. GAN primarily focuses on photo-realistic generation~\cite{radford2015unsupervised,karras2019style,zhang2021styleswin}. Other deep generative models such as variational autoencoder (VAE)~\cite{VAEICLR2014}, normalizing flows (NF)~\cite{rezende2015variational,NF_PAMI2021}, and diffusion models (DM)~\cite{ho2020denoising} are getting popular but their generation performances are less competitive than large GAN models~\cite{karras2019style,zhang2021styleswin}. Recent advances in energy-based models (EBM) show that EBM works well with supervised classification~\cite{JEM_ICLR2020,VERA_ICLR2021}, and the line of latent space EBM~\cite{pangNIPS2020,pangICML2021,zhangNIPS2021,xiaoICLR2021,nijkampICLR2022} could further stand on the power of GAN and VAE to achieve superior generative and discriminative performance. Moreover, the learned energy is found effective to detect the unknown samples from out-of-distribution data~\cite{liu2020energy,wang2021can,Lin_2021_CVPR,wang2021energy}. In this paper, we aim to exploit the latent space EBM for the fine-grained OSR problem, which has never been explored in the literature.

\section{Proposed Method} 
\label{sec:method}

\paragraph{Overview}

To achieve the fine-grained open-set recognition by latent space EBM, we introduce a latent variable $\mathbf{z}$, and formulate the joint distribution of input image $\mathbf{x}$, class label $\mathbf{y}$, attribute label $\mathbf{a}$, and $\mathbf{z}$ in a full probabilistic model:
\begin{equation}
    p(\mathbf{x},\mathbf{z},\mathbf{a},\mathbf{y}) = p_{\alpha}(\mathbf{y}|\mathbf{a},\mathbf{z})p_{\omega}(\mathbf{a}|\mathbf{z})p_{\alpha}(\mathbf{z})p_{\beta}(\mathbf{x}|\mathbf{z})
\label{eq:model}
\end{equation}
where the class label $\mathbf{y}\in\mathbb{R}^K$ is dependent on attribute $\mathbf{a}\in\mathbb{R}^M$ given the latent feature $\mathbf{z}\in \mathbb{R}^d$ such that $p(\mathbf{y},\mathbf{a}|\mathbf{z})=p_{\alpha}(\mathbf{y}|\mathbf{a},\mathbf{z})p_{\omega}(\mathbf{a}|\mathbf{z})$. The third term $p_{\alpha}(\mathbf{z})$ is an energy prior distribution of $\mathbf{z}$ in the latent space. The last term $p_{\beta}(\mathbf{x}|\mathbf{z})$ is the generative model of data $\mathbf{x}$. 

\begin{figure}[t]
    \centering
    \subcaptionbox{SVEBM~\cite{pangICML2021}\label{fig:svebm}}{
        \includegraphics[width=0.45\linewidth]{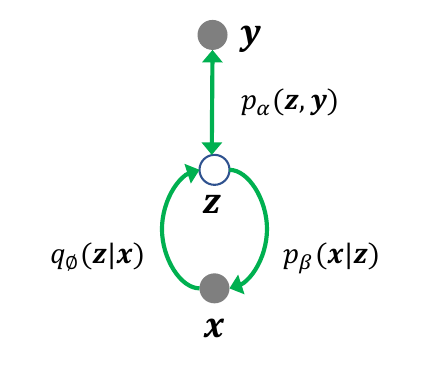}
    }
    \hspace{-4mm}
    \subcaptionbox{Ours \label{fig:fullmodel}}{
        \includegraphics[width=0.35\textwidth]{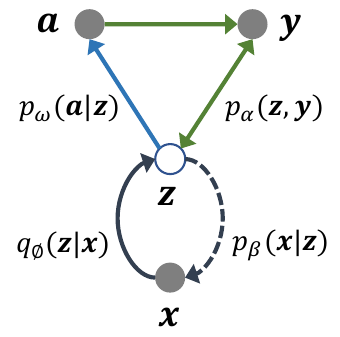}
    }
    \caption{\small{\textbf{SVEBM and the proposed model.} Compared to SVEBM, our model introduces attribute prediction to address the class granularity of the fine-grained OSR problem.}}
\end{figure}

\begin{figure*}
    \centering
    \includegraphics[width=0.95\linewidth]{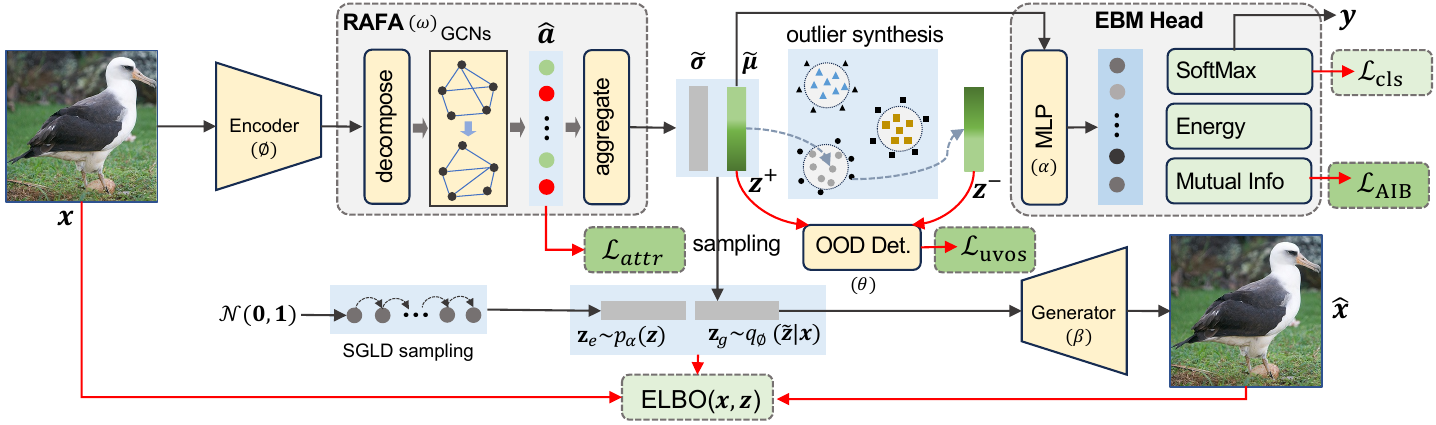}
    \caption{\textbf{Training Pipeline.} Given the training data $\mathbf{x}$, the visual encoder $\boldsymbol{\phi}$ first extracts features for the proposed RAFA module to predict the variational posterior $\mathcal{N}(\tilde{\boldsymbol{\mu}},\tilde{\boldsymbol{\sigma}})$. Then, the EBM head uses $\tilde{\boldsymbol{\mu}}$ for classification and the generator uses the sampled $\mathbf{z}_g$ (or $\mathbf{z}_e$) for image generation. The model is learned by optimizing the ELBO, $\mathcal{L}_{\text{cls}}$, $\mathcal{L}_{\text{AIB}}$, $\mathcal{L}_{\text{attr}}$, and $\mathcal{L}_{\text{uvos}}$. Red arrows indicate the training processes.}
    \label{fig:train}
\end{figure*}

Based on the SVEBM~\cite{pangICML2021} (Fig.~\ref{fig:svebm}), Fig.~\ref{fig:fullmodel} illustrates the proposed probabilistic formulation, which shows that the encoder $\boldsymbol{\phi}$, the attribute classifier $\boldsymbol{\omega}$, the open-set classifier $\boldsymbol{\alpha}$, and the generator $\boldsymbol{\beta}$ are jointly learned in an end-to-end manner. Our goal is to maximize the log-likelihood of Eq.~\eqref{eq:model} so that the discriminative open-set classification can be achieved by $p_{\alpha}(\mathbf{y}|\mathbf{a},\mathbf{z})$ while the image generation can be fulfilled by $p_{\beta}(\mathbf{x}|\mathbf{z})$ given the sampled latent $\mathbf{z}$. A complete overview of our model training pipeline is shown in Fig.~\ref{fig:train}, which will be introduced in the following sections.



\subsection{Preliminary: Latent Space EBM}
\label{sec:preliminary}

To maximize the log-likelihood of Eq.~\eqref{eq:model}, we have $\log p(\mathbf{x},\mathbf{z},\mathbf{a},\mathbf{y})=\log p(\mathbf{x},\mathbf{z},\mathbf{y}|\mathbf{a}) + \log p_{\omega}(\mathbf{a}|\mathbf{z})$ according to the implied conditional independency (see Appendix A for detailed derivation). With the energy prior of the latent variable $\mathbf{z}$, the first term $\log p(\mathbf{x},\mathbf{z},\mathbf{y}|\mathbf{a})$ is essentially a latent space EBM~\cite{pangICML2021} conditioned on the predicted attribute $\mathbf{a}$. Therefore, in this section, we primarily present the preliminary of the latent space EBM, termed as SVEBM.

Formally, SVEBM assumes the generative model to be $p(\mathbf{y}, \mathbf{z}, \mathbf{x})=p_{\alpha}(\mathbf{y}|\mathbf{z})p_{\alpha}(\mathbf{z})p_{\beta}(\mathbf{x}|\mathbf{z})$. The first two terms $p_{\alpha}(\mathbf{y}|\mathbf{z})p_{\alpha}(\mathbf{z})$ couple the multi-class label $\mathbf{y}\in \mathbb{R}^K$ and the continuous vector $\mathbf{z}\in \mathbb{R}^d$ in the $d$-dimensional latent space, 
and the EBM prior $p_{\alpha}(\mathbf{z})$ is defined as the correction of a non-informative Gaussian $p_0(\mathbf{z})$:
\begin{equation}
    p_{\alpha}(\mathbf{z}) = \frac{1}{Z_{\alpha}}\exp\left(-E_{\alpha}(\mathbf{z})\right)p_{0}(\mathbf{z}),
\end{equation}
where $Z_{\alpha}$ is the partition function and $E_{\alpha}(\mathbf{z})$ is the energy function computed by \emph{LogSumExp}, \ie~$E_{\alpha}(\mathbf{z})=-\log\sum_{k=1}^{K}\exp(f_{\alpha}(\mathbf{z})[k])$. 
Due to the intractability of $Z_{\alpha}$, directly maximizing the joint likelihood $p(\mathbf{y}, \mathbf{z}, \mathbf{x})$ w.r.t. parameters $\Theta=(\alpha,\beta)$ is infeasible. Instead, following recent EBMs, SVEBM tackles this problem by Stochastic Gradient Langevin Dynamics (SGLD)~\cite{welling2011bayesian}, which is a MCMC sampling technique to approximate $p_{\alpha}(\mathbf{z})$. However, the sampling procedure needs samples from the posterior distribution $p_{\Theta}(\mathbf{z}|\mathbf{x})$, which is still intractable. 

Instead of a second SGLD to approximate $p_{\Theta}(\mathbf{z}|\mathbf{x})$ in~\cite{pangNIPS2020}, that requires back-propagating through a large generation model $p_{\beta}(\mathbf{x}|\mathbf{z})$, SVEBM amortizes the posterior following the VAE~\cite{VAEICLR2014} which uses a variational posterior $q_{\phi}(\mathbf{z}|\mathbf{x})$ to approximate the true posterior $p_{\Theta}(\mathbf{z}|\mathbf{x})$ by maximizing the evidence lower bound (ELBO) of $\log p_{\Theta}(\mathbf{x})$:
\begin{equation}
    \text{ELBO}(\mathbf{x}|\Theta, \phi) = \log p_{\Theta}(\mathbf{x}) - \mathbb{D}_{\text{KL}}\infdivx{q_{\phi}(\mathbf{z}|\mathbf{x})}{p_{\Theta}(\mathbf{z}|\mathbf{x})}
\end{equation}
Eventually, the prior $p_{\alpha}(\mathbf{z})$, variational posterior $q_{\phi}(\mathbf{z}|\mathbf{x})$, and generator $p_{\beta}(\mathbf{x}|\mathbf{z})$ are updated by taking the gradients from the ELBO objective w.r.t. model parameters $(\alpha,\phi,\beta)$, inducing the EBM training loss:
\begin{equation}
    \mathcal{L}_{\text{EBM}}(\alpha) =  \mathbb{E}_{q_{\phi}(\mathbf{z}|\mathbf{x})}\left[\nabla_{\alpha}E_{\alpha}(\mathbf{z})\right] -\mathbb{E}_{p_{\alpha}(\mathbf{z})}\left[\nabla_{\alpha}E_{\alpha}(\mathbf{\mathbf{z}})\right]
\label{eq:elbo_ebm}
\end{equation}
and the energy-based VAE loss:
\begin{equation}
\begin{split}
    \mathcal{L}_{\text{E-VAE}}(\beta, \phi) &=  -\mathbb{E}_{q_{\phi}(\mathbf{z}|\mathbf{x})}\left[\log p_{\beta}(\mathbf{x}|\mathbf{z})\right] + \mathbb{E}_{q_{\phi}(\mathbf{z}|\mathbf{x})}[E_{\alpha}(\mathbf{z})] \\
    & + \mathbb{D}_{\text{KL}}\infdivx{q_{\phi}(\mathbf{z}|\mathbf{x})}{p_0(\mathbf{z})},
\label{eq:elbo_evae}
\end{split}
\end{equation}
where sampling from prior $p_{\alpha}(\mathbf{z})$ and variational posterior $q_{\phi}(\mathbf{z}|\mathbf{x})$ are achieved by the SGLD and the reparametrization of encoder output $f_{\phi}(\mathbf{x})$, respectively. The rest parts in Eq.~\eqref{eq:elbo_ebm} and~\eqref{eq:elbo_evae} are all computationally tractable.

For all samples $\mathbf{x}$ from data distribution $q_{\text{data}}(\mathbf{x})$, the expectation $\mathbb{E}_{q_{\text{data}}(\mathbf{x})}[\text{ELBO}(\mathbf{x}|\Theta, \phi)]$ indicates a mutual information $\mathcal{I}(\mathbf{x},\mathbf{z})$ to be minimized in the KL divergence term $\mathbb{D}_{\text{KL}}\infdivx{q_{\phi}(\mathbf{x},\mathbf{z})}{p_{\Theta}(\mathbf{x},\mathbf{z})}$. Thus, to enable classification, SVEBM proposes to add a mutual information $\mathcal{I}(\mathbf{z},\mathbf{y})$, resulting in an information bottleneck (IB) term in the loss:
\begin{equation}
\begin{split}
    \mathcal{L}_{\text{SVEBM}}(\alpha,\beta,\phi) & = -\mathcal{H}(\mathbf{x}) - \mathbb{E}_{Q_{\phi}(\mathbf{x},\mathbf{z})}\left[\log p_{\beta}(\mathbf{x}|\mathbf{z})\right] \\
    & + \mathbb{D}_{\text{KL}}\infdivx{q_{\phi}(\mathbf{z})}{p_{\alpha}(\mathbf{z})} + \mathcal{IB}(\mathbf{x},\mathbf{z},\mathbf{y})
\end{split}
\end{equation}
where the $\mathcal{IB}(\mathbf{x},\mathbf{z},\mathbf{y})=\mathcal{I}(\mathbf{x},\mathbf{z})-\lambda \mathcal{I}(\mathbf{z},\mathbf{y})$ is the IB where the weight $\lambda$ controls the compressivity of $\mathbf{z}$ over the data $\mathbf{x}$ and the expressivity of $\mathbf{z}$ to the label $\mathbf{y}$. $\mathcal{H}(x)$ is the constant data entropy, $q_{\phi}(\mathbf{z})=\mathbb{E}_{q_{\text{data}}(\mathbf{x})}[q_{\phi}(\mathbf{z}|\mathbf{x})]$. 

In summary, SVEBM aims to optimize the $\log p(\mathbf{y},\mathbf{z},\mathbf{x})$ that consists of the classification loss $\mathcal{L}_{\text{cls}}$ and $\text{ELBO}(\mathbf{x},\mathbf{z})$, along with $\mathcal{I}(\mathbf{z},\mathbf{y})$. Here, $\mathcal{L}_{\text{cls}}=-\mathbb{E}_{q_{\phi_1}(\mathbf{z}|\mathbf{x})}[\log p_{\alpha}(\mathbf{y}|\mathbf{z})]$ is a multi-class cross-entropy loss, and the $\text{ELBO}(\mathbf{x},\mathbf{z})$ is determined by Eq.~\eqref{eq:elbo_ebm} and ~\eqref{eq:elbo_evae}.

\subsection{Attribute-aware Information Bottleneck}

\paragraph{Motivation} For the FineOSR problem, the expressivity of latent variables $\mathbf{z}$ w.r.t. target class is even more important than traditional OSR, when optimizing the information bottleneck (IB) objective. Motivated by the fact that image attributes provide expressive features to identify the fine-grained classes, we propose to condition the IB over the attributes of each target class.


Specifically, for the second IB term $\mathcal{I}(\mathbf{z}, \mathbf{y})$, according to the definition of MI and the introduced attribution condition, we have the attribute-aware IB (AIB) objective:
\begin{equation}
\begin{split}
    & \mathcal{I}(\mathbf{z}, \mathbf{y}) = \mathcal{H}(\mathbf{y}) - \mathcal{H}(\mathbf{y}|\mathbf{a},\mathbf{z}) \\ &= \mathcal{H}\left(\mathbb{E}_{q_{\phi}(\mathbf{z})}[p_{\alpha}(\mathbf{y}|\mathbf{z},\mathbf{a})]\right) - \mathbb{E}_{q_{\phi}(\mathbf{z})}[\mathcal{H}(p_{\alpha}(\mathbf{y}|\mathbf{z},\mathbf{a}))],
\end{split}
\end{equation}
and the entropy $\mathcal{H}$ over $\mathbf{p}\in \mathbb{R}^K$ is given by $\mathcal{H}(\mathbf{p})=-\sum_{k=1}^K p_k\log p_k$, it is clear that MI measures the model uncertainty using the predictive distribution $p_{\alpha}(\mathbf{y}|\mathbf{z},\mathbf{a})$ over $K$ classes given the attributes $\mathbf{a}$ and sampled $\mathbf{z}$. 

Moreover, for supervised classification problems like the OSR, without explicit class supervision, the expressivity of $\mathbf{z}$ will be limited in training. To this end, we propose to couple the class label $\mathbf{y}$ with the MI by an inner-product between $\mathbf{y}$ and $\mathcal{H}(\mathbf{p})$, \ie, $\mathcal{CH}(\mathbf{p})= \langle \mathbf{y}, -\mathbf{p}\log\mathbf{p} \rangle$. 
In this way, the expressivity of latent $\mathbf{z}$ can be enhanced by optimizing the proposed attribute-aware IB objective.

\textbf{Discussion.} Since the class label $\mathbf{y}$ is one-hot where we assume $y_k=1$, the class-dependent entropy intrinsically reduces to a weighted cross-entropy given by $\mathcal{CH}(\mathbf{p})=-p_k\log p_k$ where the weight $p_k$ is the confidence value on class $k$. Thus, the merits of cross-entropy supervision are taken into account in the training.

\subsection{Residual Attribute Feature Aggregation}

\paragraph{Motivation} For the FineOSR problem, the \emph{granularity} is motivated by the fact that more fine-grained features can better identify different known classes and identify the unknown, and the fine-grained features can be identified by finding which attribute feature is important to the category. Therefore, we propose the Residual Attribute Feature Aggregation (RAFA) module to aggregate attribute-relevant features for latent variable representation. 

\begin{figure}[t]
    \centering
    \includegraphics[width=0.95\linewidth]{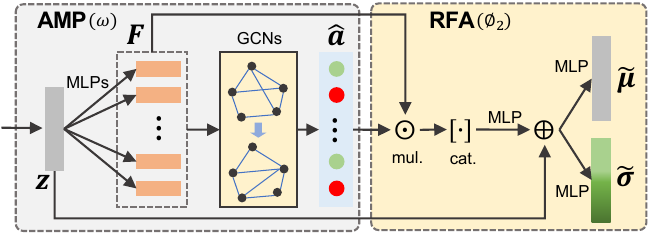}
    \caption{\textbf{Residual Attribute Feature Aggregation (RAFA)}. The image-level encoded feature $\mathbf{z}$ is first decomposed as the attribute-relevant features $\mathbf{F}\in\mathbb{R}^{M\times D}$ by $M$ MLPs. Then, GCNs are applied to predict the attribute scores $\mathbf{\hat{a}}$, which is further utilized to aggregate $\mathbf{F}$ for predicting the mean $\boldsymbol{\tilde{\mu}}$ and variance $\boldsymbol{\tilde{\sigma}}$ of the variational posterior distribution of $\tilde{\mathbf{z}}$.}
    \label{fig:ramp}
\end{figure}

Specifically, the RAFA module consists of an attribute mask prediction (AMP) module $\omega$ and a residual feature aggregation (RFA) $\phi_2$ as shown in Fig.~\ref{fig:ramp}.


\vspace{-4mm}
\paragraph{Attribute Mask Prediction (AMP)} Given an image $\mathbf{x}\in \mathbb{R}^{3\times H\times W}$, the global image feature $\mathbf{z}\in \mathbb{R}^D$ is obtained by $\mathbf{z}=f_{\phi_1}(\mathbf{x})$. The feature extractor $f_{\phi_1}$ is instantiated by ResNet~\cite{he2016deep} or ViT~\cite{vitICLR2021}. To enhance $\mathbf{z}$ with aware of fine-grained attributes, our AMP module first decomposes $\mathbf{z}$ into $M$ attribute-relevant features $\{\mathbf{f}_1,\ldots,\mathbf{f}_M\}$ where $\mathbf{f}_m\in\mathbb{R}^d$ by $M$ MLP layers $\mathbf{f}_i=h_i(\mathbf{z})$. Then, we follow the recent graph-based multi-label classification (MLC) literature~\cite{chen2019multi,Chen_2019_ICCV,ChenGCNPAMI2021} and propose to use GCN~\cite{GCNICLR2017} to predict the multi-hot attribute mask, since the GCN-based method could handle the challenges of attribute dependency and incompleteness from data annotations. 

Specifically, the $M$ features are structured as graph nodes $\mathbf{F}=[\mathbf{f}_1,\ldots,\mathbf{f}_M]^T$ where $\mathbf{F}\in \mathbb{R}^{M\times d}$. Similar to~\cite{ChenGCNPAMI2021}, the adjacent matrix $\mathbf{A}\in \mathbb{R}^{M\times M}$ indicating the correlation between attributes is obtained from closed-set training data. With the graph data $\boldsymbol{\mathcal{G}}=(\mathbf{F},\mathbf{A})$ as the input, the attribute scores which indicate the presence of attributes are predicted by $\hat{\mathbf{a}}=f_{\omega}(\boldsymbol{\mathcal{G}})$ where $\hat{\mathbf{a}}\in \mathbb{R}^M$. In practice, we use two GCN layers to instantiate $f_{\omega}$. Therefore, the objective of attribute mask prediction is to minimize the negative log-likelihood $-\log p_{\omega}(\mathbf{a}|\mathbf{z})$:
\begin{equation}
    \mathcal{L}_{\text{attr}}(\phi_1,\omega) = -\mathbb{E}_{q_{\phi_1}(\mathbf{z}|\mathbf{x})}[\log p_{\omega}(\mathbf{a}|\mathbf{z})]
\label{eq:amp_loss}
\end{equation}
Here, we use the binary cross-entropy (BCE) loss, which is a common practice in MLC literature.

\vspace{-2mm}
\paragraph{Residual Feature Aggregation (RFA)} We treat the predicted attribute scores $\hat{\mathbf{a}}$ as the soft mask to aggregate the graph node features $\mathbf{F}$. The aggregation is achieved by element-wise multiplication (mul.) and concatenation (cat.) between $\mathbf{F}$ and $\hat{\mathbf{a}}$, followed by an MLP to map the feature into the same feature space as $\mathbf{z}$. Before applying MLP layers to predict the mean $\tilde{\boldsymbol{\mu}}$ and the standard deviation $\tilde{\boldsymbol{\sigma}}$ of the variational posterior, we introduce a residual connection to the input feature $\mathbf{z}$ to facilitate the training. The following equations show the above procedure:
\begin{equation}
\begin{split}
    & \tilde{\boldsymbol{\mu}},\tilde{\boldsymbol{\sigma}} = f_{\phi2}(\mathbf{z} + \text{MLP}(\text{cat}\left(\hat{\mathbf{a}} \odot \mathbf{F}\right))) \\
    & \tilde{\mathbf{z}} = \tilde{\boldsymbol{\mu}} + \mathbf{n}\odot \tilde{\boldsymbol{\sigma}}
\label{eq:ramp}
\end{split}
\end{equation}
where $f_{\phi2}$ is typically implemented by MLPs and the second equation is the reparametrization of variational posterior sampling similar to VAE~\cite{VAEICLR2014}, \ie, $\mathbf{n}\sim \mathcal{N}(\mathbf{0},\mathbf{1})$. The residual connection in Eq.~\eqref{eq:ramp} indicates that the attribute-dependent features are aggregated to predict the the fine-grained feature and the global feature $\mathbf{z}$. Note that in training, the parameters $\phi=\{\phi_1,\phi_2\}$ are learned by the ELBO loss, and the parameters $\omega$ are learned by Eq.~\eqref{eq:amp_loss} with attribute labels as supervision. 

\subsection{Uncertainty-aware Virtual Outlier Synthesis}

\paragraph{Motivation} To identify the unknown, the open space risk of FineOSR has to be managed, i.e., the unknown samples are expected to reside in low-density areas in the latent space. This motivates our emphasis on the probability \emph{density} of samples to address the challenge of FineOSR. Benefited by the generative modeling, the density estimation is feasible since the variational posterior $p_{\phi}(\mathbf{z}|\mathbf{x})$ of latent space EBM is naturally Gaussian.

Inspired by the recent virtual outlier synthesis (VOS) method~\cite{DuICLR2022}, we propose an effective and efficient VOS method for regularizing the density estimation by leveraging the variational posterior distribution in the latent space. The method consists of class-wise density estimation and virtual outlier regularization as introduced below.

\vspace{-2mm}
\paragraph{Class-wise Density Estimation}
Given the training data $\{\mathbf{x}_i\}|_{i=1}^{N^{(b)}}$ where $N^{(b)}$ is the number of samples in the $b$-th mini-batch, the variational posterior model $q_{\Phi}(\tilde{\mathbf{z}}|\mathbf{x})$ produces the sample-wise mean $\boldsymbol{\mu}_i$ and $\boldsymbol{\sigma}_i$ for each observation $\mathbf{x}_i$. Consider the i.i.d. property of the multivariate Gaussian posterior~\cite{VAEICLR2014}, we denote the precision matrix $\mathbf{P}_i=\mathbf{V}_i^{-1}$ where the variance matrix is diagonal, \ie, $\mathbf{V}_i=\text{diag}(\sigma_{i1}^2,\ldots,\sigma_{iD}^2)$. Then, for a specific class $k\in [1,\ldots,K]$ in the $b$-th batch, the probability density is the joint distribution $\prod_{i=1}^{N_{k}^{(b)}} \mathcal{N}(\boldsymbol{\mu}_i, \mathbf{P}_i)$ where the class-wise mean $\boldsymbol{\mu}_{k}^{(b)}$ and precision $\mathbf{P}_{k}^{(b)}$ can be estimated by 
\begin{equation}
    \hat{\mathbf{P}}_k^{(b)}\hat{\boldsymbol{\mu}}_k^{(b)}  = \sum_{i=1}^{N_{k}^{(b)}} \mathbf{P}_i \boldsymbol{\mu}_i, \;\;\;\;\; \hat{\mathbf{P}}_k^{(b)} = \sum_{i=1}^{N_{k}^{(b)}} \mathbf{P}_i
\end{equation}
Here, $\hat{\boldsymbol{\mu}}_k^{(b)}$ and $\hat{\mathbf{P}}_k^{(b)}$ are biased estimates for class $k$ because only a mini-batch of training data are utilized. On the entire training dataset, we propose an online sequential update:
\begin{equation}
\begin{split}
    &\hat{\mathbf{P}}_k^{(b)}\hat{\boldsymbol{\mu}}_k^{(b)}  = \hat{\mathbf{P}}_k^{(b-1)}\hat{\boldsymbol{\mu}}_k^{(b-1)} + \sum_{i=1}^{N_{k}^{(b)}} \mathbf{P}_i \boldsymbol{\mu}_i, \\
    & \hat{\mathbf{P}}_k^{(b)} = \hat{\mathbf{P}}_k^{(b-1)}  + \sum_{i=1}^{N_{k}^{(b)}} \mathbf{P}_i
\end{split}
\end{equation}
In this way, the estimated class-wise density given by $(\hat{\boldsymbol{\mu}}_k,\hat{\mathbf{P}}_k)$ after iterating over all training samples is equivalent to the unbiased estimate when the scale of training data is sufficiently large. Moreover, 
the class-wise precision matrices maintain the varieties of distributions for different fine-grained classes, which are adaptive to detecting both the easy and hard unknown samples.

\vspace{-2mm}
\paragraph{Virtual Outlier Regularization} With the estimated class-wise Gaussian density $\mathcal{N}(\hat{\boldsymbol{\mu}}_k,\hat{\mathbf{P}}_k)$, the virtual outliers are sampled and normalized from the $\epsilon$-likelihood region:
\begin{equation}
    \mathcal{V}_k = \left\{(\mathbf{z} - \hat{\boldsymbol{\mu}}_k )\sqrt{\hat{\mathbf{P}}_k} \middle|\mathcal{N}(\mathbf{z};\hat{\boldsymbol{\mu}}_k,\hat{\mathbf{P}}_k) < \epsilon \right\}
\label{eq:uvos}
\end{equation}
where the $\epsilon$ is sufficiently small to ensure that the sampled virtual outliers are close to the class-wise decision boundary. Note in Eq.~\eqref{eq:uvos}, we normalize samples within each class by $(\hat{\boldsymbol{\mu}}_k,\hat{\mathbf{P}}_k)$. This is because, for open set recognition, known classes are distributed far away from each other so that the sampled virtual outliers are collectively too diverse to differentiate from known class data.

To regularize the decision boundary of known classes, we construct the binary dataset, in which the positive (virtual unknown) data are $\mathcal{V}^+=\cup_{k=1}^K \mathcal{V}_k$ while the negative (real known) data are from $\mathcal{V}^-=\cup_{k=1}^K \left\{(\tilde{\mathbf{z}}_i - \hat{\boldsymbol{\mu}}_k )\sqrt{\hat{\mathbf{P}}_k}\right\}\big|_{i=1}^{N^{(b)}}$. As shown in Fig.~\ref{fig:train}, we introduce a small subnetwork (OOD Det.) to differentiate between the positive and negative samples using the binary cross-entropy:
\begin{equation}
    \mathcal{L}_{\text{uvos}}(\phi, \theta) = -\mathbb{E}_{q_{\phi}(\tilde{\mathbf{z}}|\mathbf{x})}\left[\log p_{\theta}(\mathbf{s}|\{\mathbf{z}^+,\mathbf{z}^-\})\right],
\end{equation}
where $\mathbf{z}^+\in \mathcal{V}^+$ and $\mathbf{z}^-\in \mathcal{V}^-$, and $\mathbf{s}\in \mathbb{R}^2$ denotes the one-hot binary target.

\paragraph{Discussion.} Compared to the VOS~\cite{DuICLR2022}, our method is theoretically more accurate in density estimation because the VOS only uses a queue of samples $\{\mathbf{z}_i\}_{i=1}^{|\mathbb{Q}|}$ with the size $|\mathbb{Q}|$ while our sequential update scheme leverages the entire training set. Moreover, our method is more suitable for fine-grained OSR scenarios, because our class-specific precision $\hat{\mathbf{P}}_k$ based on sample-wise uncertainty $\boldsymbol{\sigma}_i$ maintains the varieties of distributions for different fine-grained classes, which are more adaptive to detecting both the easy and hard unknown samples, while VOS uses the deterministic $\mathbf{z}_i$ to estimate a shared covariance matrix between all classes. Lastly, our method is more memory efficient than VOS, because only the $(\hat{\boldsymbol{\mu}}_k,\hat{\mathbf{P}}_k)$ are stored and updated in the memory without maintaining a queue of samples. 

\subsection{Inference}

\paragraph{Open Set Recognition} To identify unknown samples from an open testing set, an effective out-of-distribution (OOD) score is critical to the testing performance. Existing EBM literature~\cite{JEM_ICLR2020,liu2020energy} advocates using the free energy given by $E(\mathbf{x}) = -\log\sum_{k=1}^K\exp(f(\mathbf{x})[k])$ for OOD detection, \ie, higher $E(\mathbf{x})$ score indicates more likely that the sample $\mathbf{x}$ is unknown. However, we empirically found that the maximum joint energy $E(\mathbf{x},y) = -f(\mathbf{x})[y]$ over $K$ classes is a better choice. Note that $\max_y E(\mathbf{x},y)$ is equivalent to the maximum logit, which was previously found effective by~\cite{maxlogit_icml22,vazeICLR2022}. In practice, the \texttt{logsumexp} of $E(\mathbf{x})$ is a mathematically smoothing form of $E(\mathbf{x},y)$ when $\mathbf{x}$ can be correctly classified as the class $y$. Thus, the empirically better performance of $\max_y E(\mathbf{x},y)$ indicates that it can additionally handle the case when $\mathbf{x}$ is misclassified. The misclassification and unknown in open-set recognition are recently studied in~\cite{cen2023devil} but out of our scope in this paper.

\vspace{-2mm}
\paragraph{Image Generation} To generate photo-realistic images, our model can achieve this by sampling the features in the latent space as the input of the trained generator. In practice, the sampling can be performed in three ways, \ie, 1) random sampling from $\mathcal{N}(\mathbf{0},\mathbf{1})$, 2) sampling from the variational posterior distribution $q_{\phi}(\mathbf{z}|\mathbf{x})$ by reparametrization trick given an input image $\mathbf{x}$, and 3) sampling from the energy-based prior distribution $p_{\alpha}(\mathbf{z})$ by the SGLD sampler. We empirically found that sampling from energy-based prior produces the most realistic images.

\section{Experiments}
\label{sec:experiments}

\subsection{Setup}
\vspace{-2mm}
\paragraph{Datasets} For the fine-grained visual recognition task, Caltech-UCSD Birds (CUB)~\cite{CUB200} is one of the most widely-used datasets that contain attribute annotations. We use its open-set split from~\cite{vazeICLR2022} that contains 100 closed-set classes and 100 unknown testing classes. The unknown classes are divided into \emph{Easy}, \emph{Medium}, and \emph{Hard} according to their attribute similarities to the training dataset. 
For the general open set recognition task, our method is implemented without the RAFA module, and compared with existing methods that are benchmarked on the TinyImageNet~\cite{le2015tiny}. The open-set split comes from~\cite{vazeICLR2022}, which contains 20 closed-set training classes and 180 unknown classes. 
For fine-grained visual generation task, in addition to the CUB dataset, we also train our model on a more large-scale cloth image dataset DeepFashion~\cite{DeepFashion}.

\paragraph{Evaluation Protocols} Following the most recent OSR literature~\cite{chenPAMI2021,vazeICLR2022,LuAAAI2022pmal}, we use the multi-class accuracy (ACC) to evaluate closed-set performance, the area under the ROC (AUROC) to evaluate the binary OOD detection performance, and the open-set classification rate (OSCR) for multi-class open-set performance. To evaluate the quality of generated fake images, we compute the Frechet Inception Distance (FID)~\cite{FID_NIPS2017} by generating $50,000$ fake images and compare with real images from datasets. A smaller FID value indicates more realistic generated images. 

\vspace{-2mm}
\paragraph{Implementation Details} Our model is implemented by PyTorch. The backbone of variational inference model $q_{\Phi}(\tilde{\mathbf{z}}|\mathbf{x})$ is implemented by ResNet-50 (\textbf{R50})~\cite{he2016deep}. We follow the same setting as~\cite{vazeICLR2022} which uses the pre-trained weights on Places~\cite{zhou2017places} dataset by the MoCo-v2~\cite{MoCov2}. Since we are the first work to address the FineOSR task, all compared methods are re-implemented with the same R50 as the backbone. To further study the scalability of our method to Transformer architectures, we additionally implement with \textbf{ViT}~\cite{vitICLR2021} as the backbone for reference. For the ViT backbone, we develop a learnable multi-scale feature fusion based on the pre-trained ViT-B/16 model from ImageNet-21K. For the generation model $p_\beta({\mathbf{x}|\mathbf{z}})$, we finetune the pre-trained DCGAN~\cite{DCGAN} and StyleSwin~\cite{StyleSwin} on the training set for fast convergence. Similar to SVEBM~\cite{pangICML2021}, our model adopts four Adam optimizers to train all learnable modules, and the learning rate is scheduled by warm-up with cosine restart as suggested by~\cite{vazeICLR2022}. We train all models for 600 epochs on CUB and 400 epochs on TinyImageNet. With one RTX 3090 GPU, our model with the ViT backbone takes 9 hours for 600 epochs of training and 54 seconds for testing. More details are in the Appendix B.

\subsection{Main Results}

\begin{table}[t]
\caption{\small Fine-grained OSR results (\%) on \textbf{CUB}~\cite{CUB200} dataset. AUROC and OSCR metrics are shown on \emph{Easy} / \emph{Medium} / \emph{Hard} splits. Here we denote JEM-$x$ and JEM-$z$ as the JEM versions that MCMC sampling in data space and feature space, respectively. All results of compared methods are re-implemented.}
\label{tab:cub_results}
    \centering
    \small
    \setlength{\tabcolsep}{1.mm}
    \setlength{\extrarowheight}{0.5mm}
    \begin{tabular}{lccc}\toprule
    Method & ACC & AUROC &OSCR \\\hline
    JEM-$x$~\cite{JEM_ICLR2020} & 34.19 & 57.94 / 56.85 / 53.65 & 25.69 / 25.31 / 24.57 \\
    JEM-$z$~\cite{JEM_ICLR2020}  & 86.17 & 85.59 / 81.82 / 76.69 & 78.02 / 75.32 / 71.30  \\
    SVEBM~\cite{pangICML2021}  & 83.01 & 80.64 / 77.56 / 73.75 & 72.03 / 69.80 / 67.29 \\
    OpenGAN~\cite{kong2021opengan} & 87.31 & 41.69 / 41.01 / 33.94 & 35.03 / 34.50 / 28.51 \\
    \hline
    Ours (R50) & 84.33 & 87.28 / 79.40 / 71.96 & 77.72 / 72.04 / 66.04  \\
    Ours (ViT) & \textbf{93.55} & \textbf{88.49} / \textbf{85.58} / \textbf{81.10} & \textbf{84.11} / \textbf{81.91} / \textbf{78.20}  \\
    \bottomrule
    \end{tabular}
\end{table}

\begin{table}[t]
\caption{\small General OSR results (\%) on \textbf{TinyImageNet} dataset. Results of compared methods are reported from published papers.}
\label{tab:tin_results}
    \centering
    \small
    \setlength{\tabcolsep}{3mm}
        \begin{tabular}{lccc}\toprule
        Method & ACC & AUROC &OSCR \\\hline
        Softmax  &  72.9  &  57.7  & 60.8   \\
        C2AE~\cite{oza2019c2ae}     &  --  &  74.8  &  --  \\
        OpenHybrid~\cite{zhang2020hybrid}  & --   &  79.3  &    \\
        RPL~\cite{chen2020learning}  &  --  & 68.8   &  53.2  \\
        ARPL~\cite{chenPAMI2021}   &  65.9  & 78.2   &  65.9   \\
        AMPF++~\cite{xia2021adversarial}  & 81.1   &  79.7  &  69.0  \\
        PMAL~\cite{lu2022pmal}  &  84.7  &  83.1  &  --  \\
        \hline
        Ours & \textbf{90.7}  &  \textbf{86.2}  & \textbf{81.0} \\
        \bottomrule
        \end{tabular}
\end{table}

\begin{table}
\centering
\captionsetup{font=small}
\caption{\small FID results for fine-grained image generation.}
\label{tab:im_gen}
\small
\setlength{\tabcolsep}{0.8mm}
    \begin{tabular}{lcc}\toprule
    Method & CUB  & DeepFashion \\
    \midrule
    SVEBM~\cite{pangICML2021} & 155.22  & 244.60 \\
    Ours (DCGAN) & 159.04 & 109.83\\
    Ours (StyleSwin) & \textbf{26.52}  & \textbf{30.54} \\
    \bottomrule
    \end{tabular}
\end{table}

\paragraph{Fine-grained Open Set Recognition}
Table~\ref{tab:cub_results} shows that our method achieves the best performance on the CUB dataset. When using the ResNet-50 (R50) backbone, our method achieves better AUROC and OSCR performance on \emph{Easy} and \emph{Medium} OSR splits than the baseline SVEBM, and outperforms the GAN-based OSR method OpenGAN by a large margin. Equipped with ViT encoder, the both closed-set and the open-set results are improved significantly. This observation is consistent with the conclusions from recent discriminative OSR works~\cite{vazeICLR2022,cai2022open}. Though the performance of our R50-based model shows some inferior FineOSR results compared to JEM-$z$, our method can additionally work well for image generation. 

\paragraph{General Open Set Recognition} Though our model is developed for the fine-grained OSR task, it can be implemented for the general OSR task without using the RAFA module. We report the results of our R50-based model on the widely-used TinyImageNet dataset as shown in Table~\ref{tab:tin_results}. We could see the clear advantages of our method over the existing discriminative OSR models. When the ViT is utilized, both the closed and open set performance are significantly improved.

\vspace{-2mm}
\paragraph{Fine-grained Visual Generation}

\begin{figure*}[t]
    \centering
    \subcaptionbox{Real: $\mathbf{x}\sim p_{\text{data}}(\mathbf{x})$\label{supfig:cub_real}}{
        \includegraphics[width=0.245\linewidth]{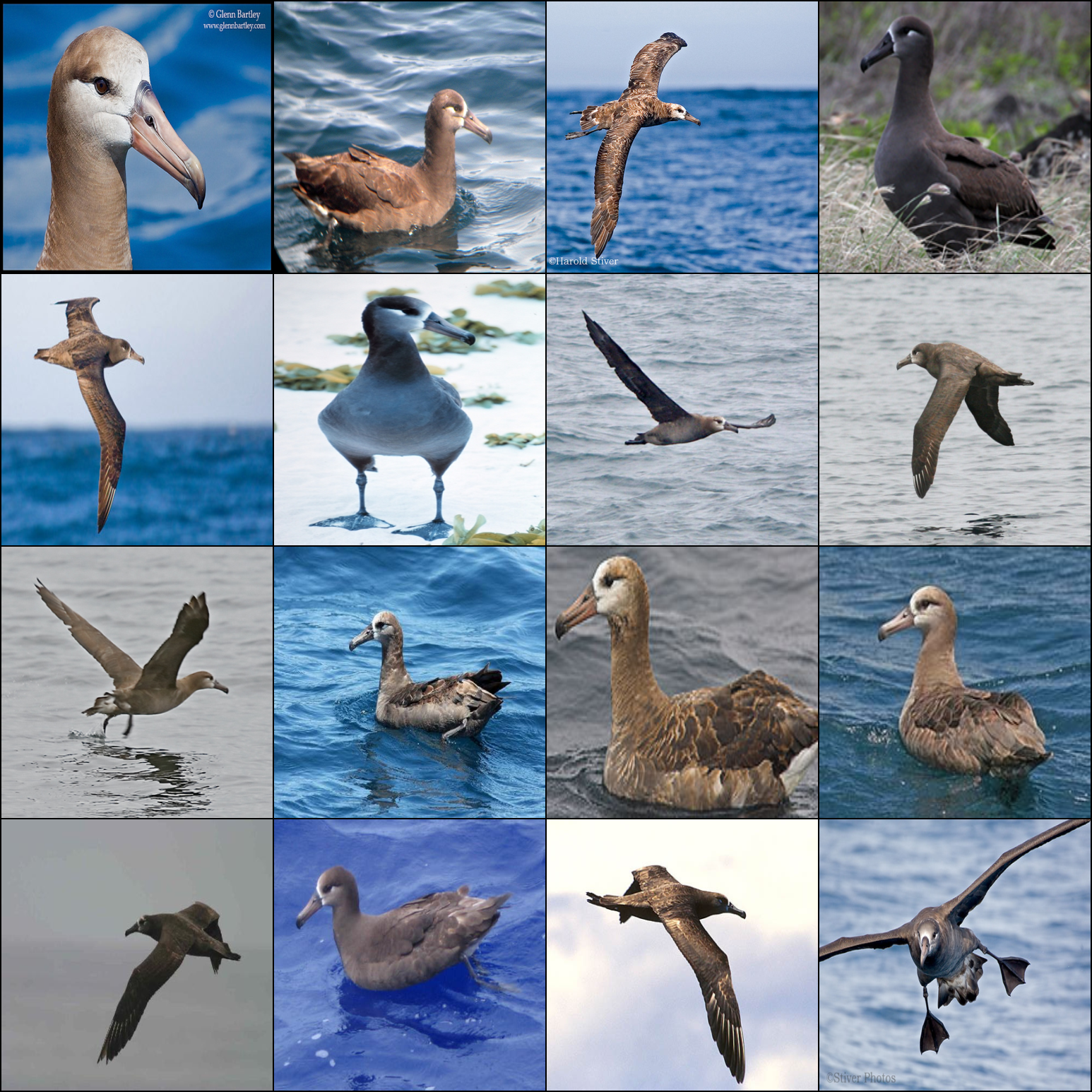}
    }
    \hspace{-3mm}
    \subcaptionbox{Fake (random): $\mathbf{z}\!\sim\! \mathcal{N}(\mathbf{0},\mathbf{1})$\label{supfig:cub_pretrain}}{
        \includegraphics[width=0.245\linewidth]{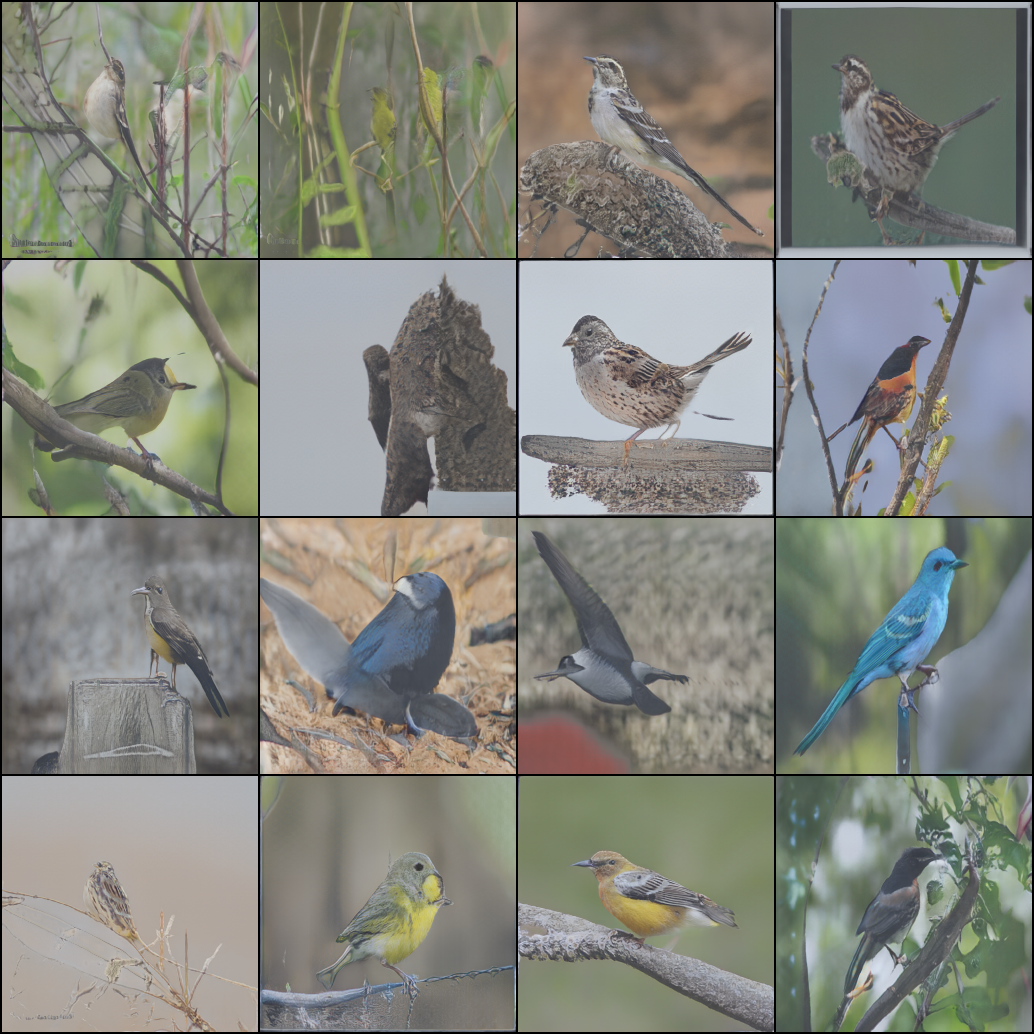}
    }
    \hspace{-3mm}
    \subcaptionbox{Fake (posterior): $\mathbf{z}\!\sim\!q_{\phi}(\mathbf{z}|\mathbf{x})$ \label{supfig:cub_post}}{
        \includegraphics[width=0.245\linewidth]{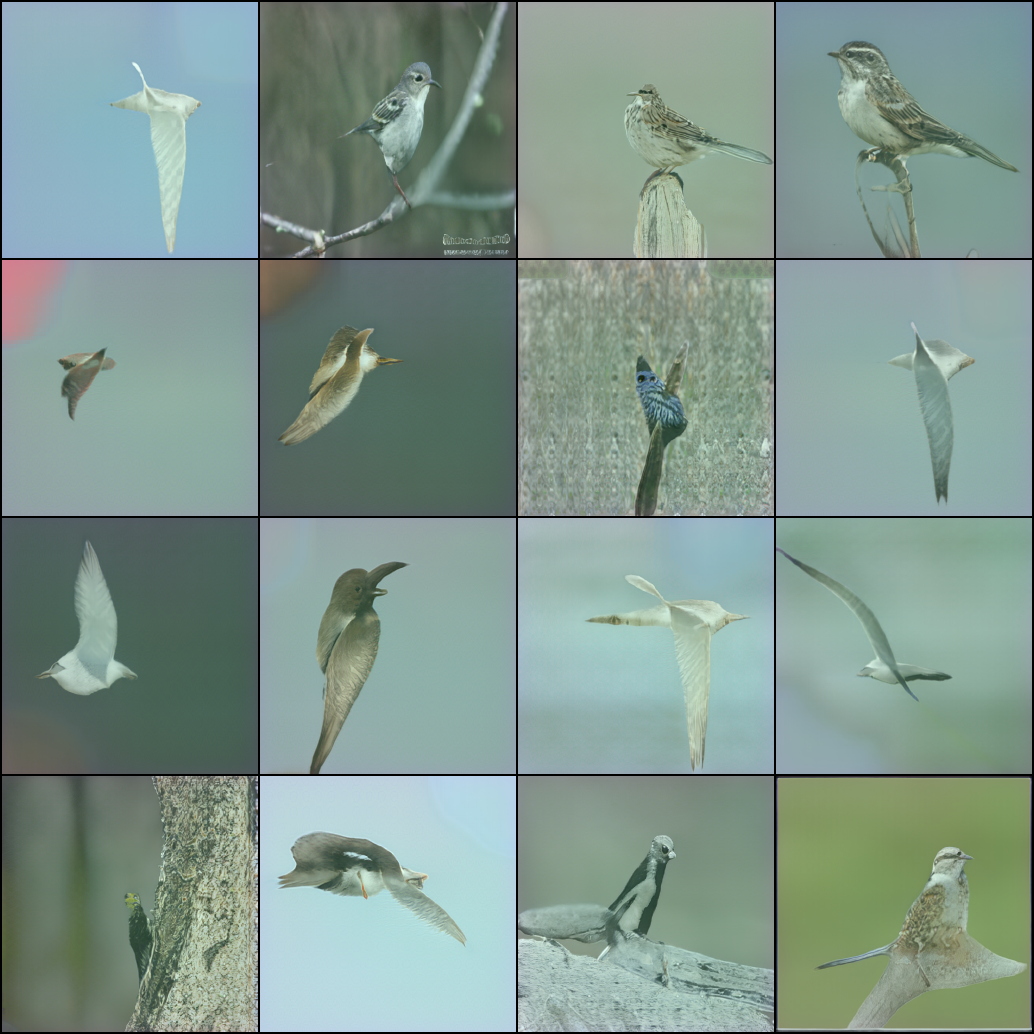}
    }
    \hspace{-3mm}
    \subcaptionbox{Fake (prior): $\mathbf{z}\!\sim\!p_{\alpha}(\mathbf{\mathbf{z}})$\label{supfig:cub_prior}}{
        \includegraphics[width=0.245\linewidth]{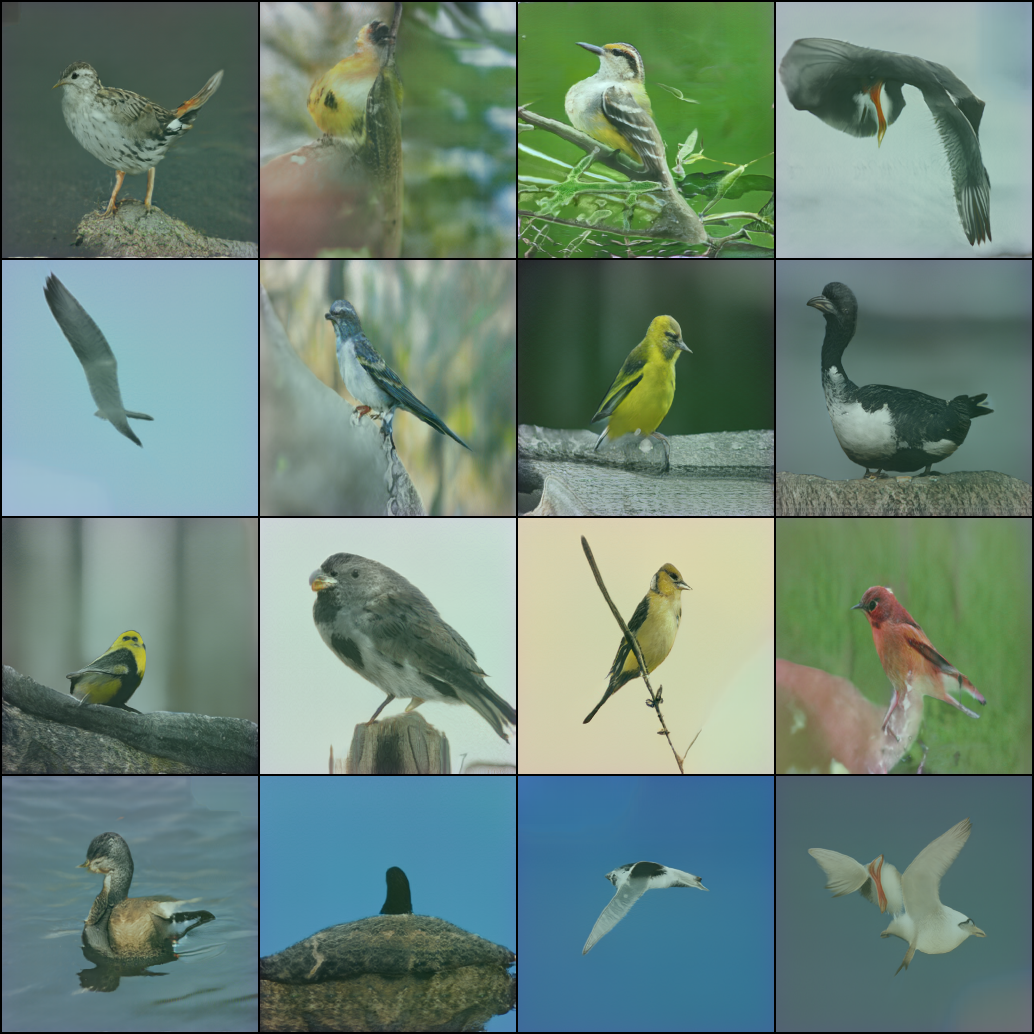}
    }
\captionsetup{font=small,aboveskip=5pt}
  \caption{\textbf{Image Generation model trained on CUB dataset}. All images are with the size $256\times 256$. We compare different methods including the random sampling as input for the pre-trained generator (Fig.~\ref{supfig:cub_pretrain}), variational posterior sampling (Fig.~\ref{supfig:cub_post}), and EBM prior sampling (Fig.~\ref{supfig:cub_prior}). Results show that the prior sampling results are the most photo-realistic.}
  \label{fig:vis_gen}
\end{figure*}

In Table~\ref{tab:im_gen}, we report the FID metric for evaluating the quality of generated images. They show that the quality of generated images by our model can be improved when utilizing the StylwSwin generator, which is a state-of-the-art ViT-based generator. When using the same DCGAN generator, our method is still better than the baseline SVEBM on the DeepFashion dataset and comparable FID scores on the CUB dataset. Besides, part of the generated images by sampling from the EBM prior $p_{\alpha}(\mathbf{z})$ are visualized in Fig.~\ref{fig:vis_gen}. Our model could generate photo-realistic images with high resolution ($256\times 256$), which is practically nontrivial for JEM to perform MCMC sampling in such a high-dimensional space.

\subsection{Model Analysis}

\paragraph{Ablation Study} 

\begin{table}[]
\centering
\captionsetup{font=small}
\caption{\textbf{Ablation study}. Results (\%) of \emph{Medium} unknown on CUB dataset are reported. For the `Ours (VOS)', we replace the proposed UVOS with VOS~\cite{DuICLR2022} method.}
\label{tab:ablation}
\small
    \begin{tabular}{lccc}\toprule
    Method &AUROC &AUPR &OSCR \\\midrule
    w/o. RAFA & 84.23 & 65.95 & 80.81 \\
    w/o. AIB & 85.41 & 69.98 &81.80  \\
    w/o. UVOS & 84.99 & 69.96 & 81.38 \\\midrule
    Ours (VOS~\cite{DuICLR2022})  &  85.44  & 67.69   &  81.55  \\
    Ours (full model) & \textbf{85.58} & \textbf{70.42} & \textbf{81.91}\\
    \bottomrule
    \end{tabular}
\end{table}

Table~\ref{tab:ablation} shows the results of removing each individual component of our full model with ViT encoder on CUB dataset. We can see that all these components contribute to the performance of detecting the unknown. Specifically, we observe the most significant contribution from RAFA module. When replacing the proposed UVOS module with the state-of-the-art method VOS~\cite{DuICLR2022}, we can see clear performance decreases on all metrics especially the AUPR. This suggests the superiority of the proposed UVOS to VOS for fine-grained open-set recognition.

\begin{figure}[t]
    \centering
    \caption{Distributions of OOD score $\max_y E(\mathbf{x},y)$.}\label{fig:ood_score}
    \subcaptionbox{CUB\label{supfig:ood_cub_vitdcgan}}{
        \includegraphics[width=0.46\linewidth]{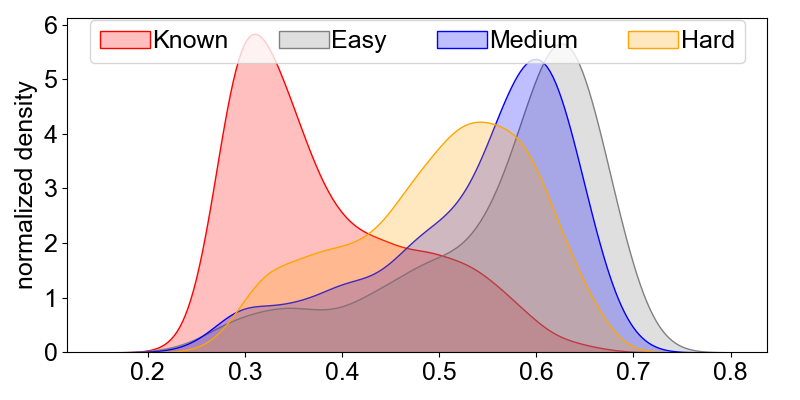}
    }
    \hspace{-2mm}
    \subcaptionbox{TinyImageNet\label{supfig:ood_tin_vitdcgan}}{
        \includegraphics[width=0.46\linewidth]{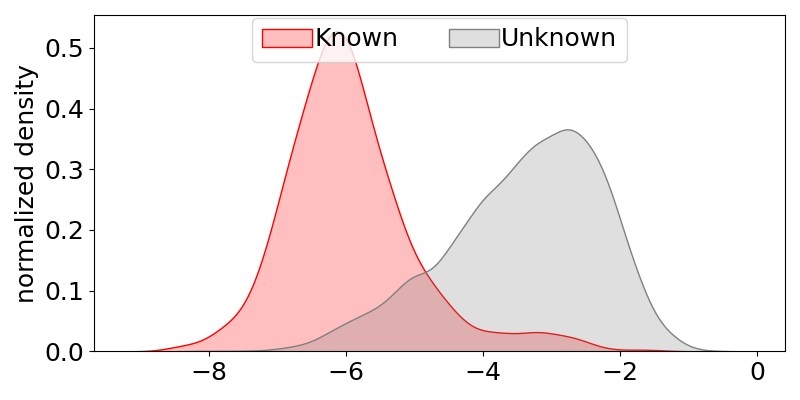}
    }
\end{figure}

\vspace{-2mm}
\paragraph{OOD Scores} In Table~\ref{tab:ood_score}, we compare the proposed OOD score by the maximum joint energy $\max_y E(\mathbf{x},y)$ with other two commonly-used OOD scores, \ie, the maximum softmax probability (denoted as $1\!\!-\!\!\max_y p(y|\mathbf{x})$) and the free energy $E(\mathbf{x})$ proposed by~\cite{liu2020energy}. The best results achieved by $\max_y E(\mathbf{x},y)$ indicate that the maximum joint energy over $K$ classes is empirically the best choice. This conclusion is consistent with the observation in recent literature~\cite{vazeICLR2022} that found the maximum logit is a strong OOD score, while our formulation by $\max_y E(\mathbf{x},y)$ provides a theoretical view of the energy distribution. Besides, Fig.~\ref{fig:ood_score} shows a clear separation between different levels of openness difficulty and the known test data by $\max_y E(\mathbf{x},y)$ on CUB dataset, as well as clear separation on the TinyImageNet dataset for identifying coarse-grained unknown samples.   

\begin{figure}[t]
    \centering
    \centering
    \subcaptionbox{SVEBM~\cite{pangICML2021}\label{fig:tsne_svebm}}{
        \includegraphics[width=0.485\linewidth]{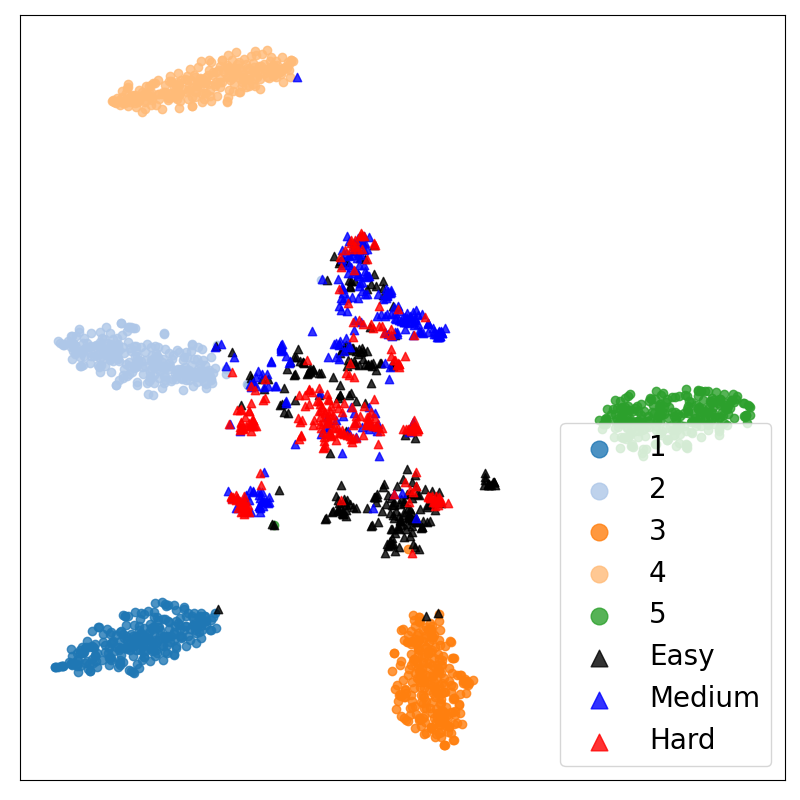}
    }
    \hspace{-4mm}
    \subcaptionbox{Ours \label{fig:tsne_ours}}{
        \includegraphics[width=0.485\linewidth]{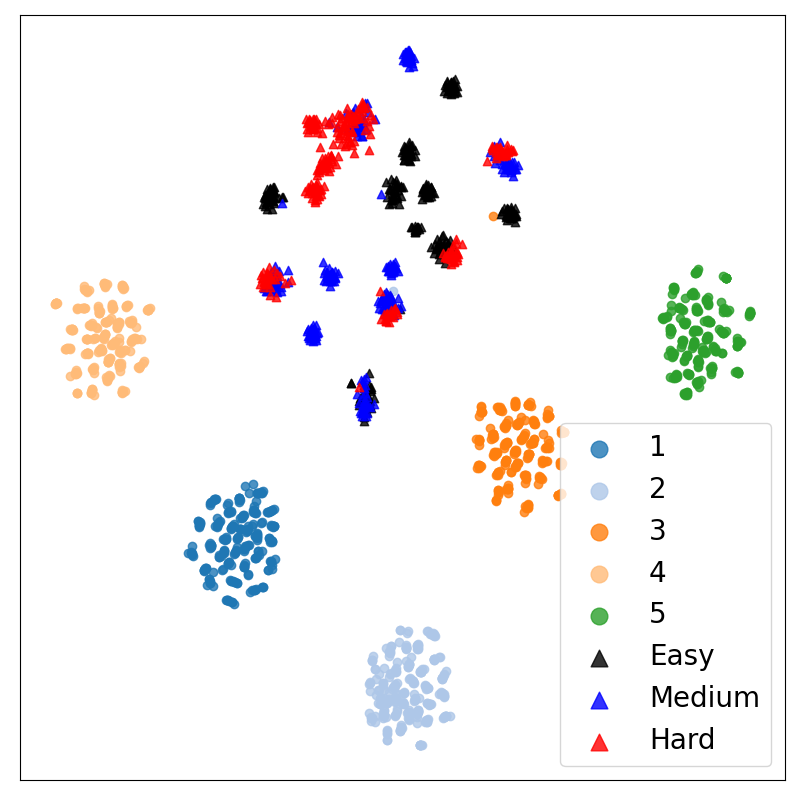}
    }
\captionsetup{font=small,aboveskip=5pt}
  \caption{\textbf{Latent feature distributions.}  We show the distribution of feature embeddings by tSNE on CUB dataset, from which 5 known classes with 256 images for each are randomly selected.}
  \label{fig:vis_embed}
\end{figure}

\vspace{-2mm}
\paragraph{Latent Feature Distributions}
In Fig.~\ref{fig:vis_embed}, the latent features visualized by tSNE show better-bounded closed- and open-set clusters of our method compared to the SVEBM baseline, especially the clustering effect on different levels of openness difficulty for fine-grained classes. More tSNE visualizations are in Appendix C.


\begin{table}
\caption{Results of using different OOD scores on CUB dataset. Numbers are reported on \emph{Easy} / \emph{Medium} / \emph{Hard} unknowns.}\label{tab:ood_score}
\centering
    \captionsetup{font=small,aboveskip=3pt}
    \small
    \setlength{\tabcolsep}{0.4mm}
    \begin{tabular}{lcc}\toprule
    Method  & AUROC (\%) & OSCR (\%) \\\midrule
    $1-\max_y p(y|\mathbf{x})$  & 88.25 / 85.32 / 80.99 & 83.94 / 81.76 / 78.15  \\
    $E(\mathbf{x})$~\cite{liu2020energy}  & \textbf{88.53} / \textbf{85.81} / 80.74 & 84.06 / 81.89 / 77.72  \\
    $\max_yE(\mathbf{x}, y)$  & 88.49 / 85.58 / \textbf{81.10}  & \textbf{84.11} / \textbf{81.91} / \textbf{78.20}\\
    \bottomrule
    \end{tabular}
\end{table}

\vspace{-2mm}
\paragraph{Limitations and Societal Impact}
Our method is limited by the computational cost in training due to the MCMC sampling for approximating the prior distribution of latent $\mathbf{z}$. This can be potentially improved by connecting with diffusion models~\cite{yu2022latent}. The potential negative impact is the risk of abuse for generating fake social-media content. The proposed method can also be utilized for many social goods. For example, the classifier is useful to discover new species of birds and the generator can be used to augment data for large-scale fine-grained visual recognition scenarios where annotations are limited. 


\vspace{-2mm}
\section{Conclusion}
\label{sec:conclusion}

In this paper, we propose to explore the existing latent space EBM for fine-grained open-set recognition (FineOSR). The main idea is to represent a high-resolution image by low-dimensional latent features and assume an energy distribution in the latent space that couples the latent features and class labels. To address the fundamental challenges of the fine-grained OSR task, we improve the expressivity and granularity of latent features for discriminative classification and leverages the learned class-wise density to synthesize the features of unknown class for OSR purpose. Our method is flexible to take advantage of vision transformers for both image classification and generation. It is effective when compared with other methods on both fine-grained and general visual recognition datasets.

\textbf{Acknowledgement.} Wentao Bao and Yu Kong are supported in part by the Office of Naval Research under grant number N00014-23-1-2046. The views and conclusions contained in this document are those of the authors and should not be interpreted as representing the official policies, either expressed or implied, of the Office of Naval Research or the U.S. Government.

\balance

{\small
\bibliographystyle{ieee_fullname}
\bibliography{egbib}

\begin{thebibliography}{10}\itemsep=-1pt

\bibitem{BaoICCV2021}
Wentao Bao, Qi Yu, and Yu Kong.
\newblock Evidential deep learning for open set action recognition.
\newblock In {\em ICCV}, 2021.

\bibitem{BendaleCVPR2016}
Abhijit Bendale and Terrance~E. Boult.
\newblock Towards open set deep networks.
\newblock In {\em CVPR}, 2016.

\bibitem{BlumICCV2019}
Hermann Blum, Paul-Edouard Sarlin, Juan Nieto, Roland Siegwart, and Cesar Cadena.
\newblock Fishyscapes: A benchmark for safe semantic segmentation in autonomous driving.
\newblock In {\em ICCV (Workshop)}, 2019.

\bibitem{cai2022open}
Feiyang Cai, Zhenkai Zhang, Jie Liu, and Xenofon Koutsoukos.
\newblock Open set recognition using vision transformer with an additional detection head.
\newblock {\em arXiv preprint arXiv:2203.08441}, 2022.

\bibitem{cen2023devil}
Jun Cen, Di Luan, Shiwei Zhang, Yixuan Pei, Yingya Zhang, Deli Zhao, Shaojie Shen, and Qifeng Chen.
\newblock The devil is in the wrongly-classified samples: Towards unified open-set recognition.
\newblock In {\em ICLR}, 2023.

\bibitem{chenPAMI2021}
Guangyao Chen, Peixi Peng, Xiangqian Wang, and Yonghong Tian.
\newblock Adversarial reciprocal points learning for open set recognition.
\newblock {\em IEEE TPAMI}, 2021.

\bibitem{chen2020learning}
Guangyao Chen, Limeng Qiao, Yemin Shi, Peixi Peng, Jia Li, Tiejun Huang, Shiliang Pu, and Yonghong Tian.
\newblock Learning open set network with discriminative reciprocal points.
\newblock In {\em ECCV}, 2020.

\bibitem{chen2018knowledge}
Tianshui Chen, Liang Lin, Riquan Chen, Yang Wu, and Xiaonan Luo.
\newblock Knowledge-embedded representation learning for fine-grained image recognition.
\newblock In {\em IJCAI}, 2018.

\bibitem{Chen_2019_ICCV}
Tianshui Chen, Muxin Xu, Xiaolu Hui, Hefeng Wu, and Liang Lin.
\newblock Learning semantic-specific graph representation for multi-label image recognition.
\newblock In {\em ICCV}, 2019.

\bibitem{MoCov2}
Xinlei Chen, Haoqi Fan, Ross Girshick, and Kaiming He.
\newblock Improved baselines with momentum contrastive learning.
\newblock Technical report, Facebook AI Research (FAIR), 2020.

\bibitem{ChenGCNPAMI2021}
Zhaomin Chen, Xiu-Shen Wei, Peng Wang, and Yanwen Guo.
\newblock Learning graph convolutional networks for multi-label recognition and applications.
\newblock {\em IEEE TPAMI}, pages 1--1, 2021.
\newblock Early Access.

\bibitem{chen2019multi}
Zhao-Min Chen, Xiu-Shen Wei, Peng Wang, and Yanwen Guo.
\newblock Multi-label image recognition with graph convolutional networks.
\newblock In {\em CVPR}, 2019.

\bibitem{Dai2021}
Wei Dai, Wenhui Diao, Xian Sun, Yue Zhang, Liangjin Zhao, Jun Li, and Kun Fu.
\newblock Camv: Class activation mapping value towards open set fine-grained recognition.
\newblock {\em IEEE Access}, 9:8167--8177, 2021.

\bibitem{vitICLR2021}
Alexey Dosovitskiy, Lucas Beyer, Alexander Kolesnikov, Dirk Weissenborn, Xiaohua Zhai, Thomas Unterthiner, Mostafa Dehghani, Matthias Minderer, Georg Heigold, Sylvain Gelly, et~al.
\newblock An image is worth 16x16 words: Transformers for image recognition at scale.
\newblock In {\em ICLR}, 2021.

\bibitem{DuICLR2022}
Xuefeng Du, Zhaoning Wang, Mu Cai, and Yixuan Li.
\newblock {VOS}: Learning what you don't know by virtual outlier synthesis.
\newblock In {\em ICLR}, 2022.

\bibitem{DubeyECCV2018}
Abhimanyu Dubey, Otkrist Gupta, Pei Guo, Ramesh Raskar, Ryan Farrell, and Nikhil Naik.
\newblock Pairwise confusion for fine-grained visual classification.
\newblock In {\em ECCV}, 2018.

\bibitem{dubey2018maximum}
Abhimanyu Dubey, Otkrist Gupta, Ramesh Raskar, and Nikhil Naik.
\newblock Maximum-entropy fine grained classification.
\newblock In {\em NeurIPS}, 2018.

\bibitem{VERA_ICLR2021}
David Duvenaud, Jacob Kelly, Kevin Swersky, Milad Hashemi, Mohammad Norouzi, and Will Grathwohl.
\newblock No mcmc for me: Amortized samplers for fast and stable training of energy-based models.
\newblock In {\em ICLR}, 2021.

\bibitem{fangICML2021}
Zhen Fang, Jie Lu, Anjin Liu, Feng Liu, and Guangquan Zhang.
\newblock Learning bounds for open-set learning.
\newblock In {\em ICML}, 2021.

\bibitem{fu2017look}
Jianlong Fu, Heliang Zheng, and Tao Mei.
\newblock Look closer to see better: Recurrent attention convolutional neural network for fine-grained image recognition.
\newblock In {\em CVPR}, 2017.

\bibitem{GeBMVC2017}
Zongyuan Ge, Sergey Demyanov, Zetao Chen, and Rahil Garnavi.
\newblock Generative {OpenMax} for multi-class open set classification.
\newblock In {\em BMVC}, 2017.

\bibitem{geng2020recent}
Chuanxing Geng, Sheng-jun Huang, and Songcan Chen.
\newblock Recent advances in open set recognition: A survey.
\newblock {\em IEEE TPAMI}, 43(10):3614--3631, 2020.

\bibitem{gillert2021towards}
Alexander Gillert and Uwe~Freiherr von Lukas.
\newblock Towards combined open set recognition and out-of-distribution detection for fine-grained classification.
\newblock In {\em VISIGRAPP}, 2021.

\bibitem{GAN}
Ian~J. Goodfellow, Jean Pouget-Abadie, Mehdi Mirza, Bing Xu, David Warde-Farley, Sherjil Ozair, Aaron Courville, and Yoshua Bengio.
\newblock Generative adversarial networks.
\newblock In {\em NeurIPS}, 2014.

\bibitem{JEM_ICLR2020}
Will Grathwohl, Kuan-Chieh Wang, J{\"o}rn-Henrik Jacobsen, David Duvenaud, Mohammad Norouzi, and Kevin Swersky.
\newblock Your classifier is secretly an energy based model and you should treat it like one.
\newblock In {\em ICLR}, 2020.

\bibitem{he2016deep}
Kaiming He, Xiangyu Zhang, Shaoqing Ren, and Jian Sun.
\newblock Deep residual learning for image recognition.
\newblock In {\em CVPR}, 2016.

\bibitem{he2017fine}
Xiangteng He and Yuxin Peng.
\newblock Fine-grained image classification via combining vision and language.
\newblock In {\em CVPR}, 2017.

\bibitem{maxlogit_icml22}
Dan Hendrycks, Steven Basart, Mantas Mazeika, Andy Zou, Joe Kwon, Mohammadreza Mostajabi, Jacob Steinhardt, and Dawn Song.
\newblock Scaling out-of-distribution detection for real-world settings.
\newblock In {\em ICML}, 2022.

\bibitem{FID_NIPS2017}
Martin Heusel, Hubert Ramsauer, Thomas Unterthiner, Bernhard Nessler, and Sepp Hochreiter.
\newblock Gans trained by a two time-scale update rule converge to a local nash equilibrium.
\newblock In {\em NeurIPS}, 2017.

\bibitem{ho2020denoising}
Jonathan Ho, Ajay Jain, and Pieter Abbeel.
\newblock Denoising diffusion probabilistic models.
\newblock In {\em NeurIPS}, 2020.

\bibitem{JainECCV2014}
Lalit~P Jain, Walter~J Scheirer, and Terrance~E Boult.
\newblock Multi-class open set recognition using probability of inclusion.
\newblock In {\em ECCV}, 2014.

\bibitem{ji2020attention}
Ruyi Ji, Longyin Wen, Libo Zhang, Dawei Du, Yanjun Wu, Chen Zhao, Xianglong Liu, and Feiyue Huang.
\newblock Attention convolutional binary neural tree for fine-grained visual categorization.
\newblock In {\em CVPR}, 2020.

\bibitem{junior2016specialized}
Pedro Ribeiro~Mendes J{\'u}nior, Terrance~E Boult, Jacques Wainer, and Anderson Rocha.
\newblock Open-set support vector machines.
\newblock {\em TSMC}, 2021.
\newblock Early Access.

\bibitem{karras2019style}
Tero Karras, Samuli Laine, and Timo Aila.
\newblock A style-based generator architecture for generative adversarial networks.
\newblock In {\em CVPR}, 2019.

\bibitem{VAEICLR2014}
Diederik~P Kingma and Max Welling.
\newblock Auto-encoding variational bayes.
\newblock In {\em ICLR}, 2014.

\bibitem{GCNICLR2017}
Thomas~N Kipf and Max Welling.
\newblock Semi-supervised classification with graph convolutional networks.
\newblock In {\em ICLR}, 2017.

\bibitem{NF_PAMI2021}
Ivan Kobyzev, Simon~J.D. Prince, and Marcus~A. Brubaker.
\newblock Normalizing flows: An introduction and review of current methods.
\newblock {\em IEEE TPAMI}, 43(11):3964--3979, 2021.

\bibitem{kong2021opengan}
Shu Kong and Deva Ramanan.
\newblock {OpenGAN}: Open-set recognition via open data generation.
\newblock In {\em ICCV}, 2021.

\bibitem{le2015tiny}
Ya Le and Xuan Yang.
\newblock Tiny imagenet visual recognition challenge.
\newblock {\em CS 231N}, 7(7):3, 2015.

\bibitem{lin2015deep}
Di Lin, Xiaoyong Shen, Cewu Lu, and Jiaya Jia.
\newblock Deep lac: Deep localization, alignment and classification for fine-grained recognition.
\newblock In {\em CVPR}, 2015.

\bibitem{lin2015bilinear}
Tsung-Yu Lin, Aruni RoyChowdhury, and Subhransu Maji.
\newblock Bilinear cnn models for fine-grained visual recognition.
\newblock In {\em ICCV}, 2015.

\bibitem{Lin_2015_ICCV}
Tsung-Yu Lin, Aruni RoyChowdhury, and Subhransu Maji.
\newblock Bilinear cnn models for fine-grained visual recognition.
\newblock In {\em ICCV}, 2015.

\bibitem{Lin_2021_CVPR}
Ziqian Lin, Sreya~Dutta Roy, and Yixuan Li.
\newblock Mood: Multi-level out-of-distribution detection.
\newblock In {\em CVPR}, 2021.

\bibitem{liu2020energy}
Weitang Liu, Xiaoyun Wang, John Owens, and Yixuan Li.
\newblock Energy-based out-of-distribution detection.
\newblock {\em Advances in Neural Information Processing Systems}, 33:21464--21475, 2020.

\bibitem{DeepFashion}
Ziwei Liu, Ping Luo, Shi Qiu, Xiaogang Wang, and Xiaoou Tang.
\newblock Deepfashion: Powering robust clothes recognition and retrieval with rich annotations.
\newblock In {\em CVPR}, 2016.

\bibitem{LuAAAI2022pmal}
Jing Lu, Yunxu Xu, Hao Li, Zhanzhan Cheng, and Yi Niu.
\newblock Pmal: Open set recognition via robust prototype mining.
\newblock In {\em AAAI}, 2022.

\bibitem{lu2022pmal}
Jing Lu, Yunlu Xu, Hao Li, Zhanzhan Cheng, and Yi Niu.
\newblock Pmal: Open set recognition via robust prototype mining.
\newblock In {\em AAAI}, 2022.

\bibitem{mundt2019open}
Martin Mundt, Iuliia Pliushch, Sagnik Majumder, and Visvanathan Ramesh.
\newblock Open set recognition through deep neural network uncertainty: Does out-of-distribution detection require generative classifiers?
\newblock In {\em ICCV (Workshop)}, 2019.

\bibitem{nijkampICLR2022}
Erik Nijkamp, Ruiqi Gao, Pavel Sountsov, Srinivas Vasudevan, Bo Pang, Song-Chun Zhu, and Ying~Nian Wu.
\newblock Mcmc should mix: Learning energy-based model with neural transport latent space mcmc.
\newblock In {\em ICLR}, 2022.

\bibitem{oza2019c2ae}
Poojan Oza and Vishal~M Patel.
\newblock C2ae: Class conditioned auto-encoder for open-set recognition.
\newblock In {\em CVPR}, 2019.

\bibitem{pangNIPS2020}
Bo Pang, Tian Han, Erik Nijkamp, Song-Chun Zhu, and Ying~Nian Wu.
\newblock Learning latent space energy-based prior model.
\newblock In {\em NeurIPS}, 2020.

\bibitem{pangICML2021}
Bo Pang and Ying~Nian Wu.
\newblock Latent space energy-based model of symbol-vector coupling for text generation and classification.
\newblock In {\em ICML}, 2021.

\bibitem{PereraCVPR2020}
Pramuditha Perera, Vlad~I Morariu, Rajiv Jain, Varun Manjunatha, Curtis Wigington, Vicente Ordonez, and Vishal~M Patel.
\newblock Generative-discriminative feature representations for open-set recognition.
\newblock In {\em CVPR}, 2020.

\bibitem{perera2019deep}
Pramuditha Perera and Vishal~M Patel.
\newblock Deep transfer learning for multiple class novelty detection.
\newblock In {\em CVPR}, 2019.

\bibitem{PrabhuNIPS2019}
Viraj Prabhu, Anitha Kannan, Geoffrey~J. Tso, Namit Katariya, Manish Chablani, David Sontag, and Xavier Amatriain.
\newblock Open set medical diagnosis.
\newblock In {\em NeurIPS (Workshop)}, 2019.

\bibitem{radford2015unsupervised}
Alec Radford, Luke Metz, and Soumith Chintala.
\newblock Unsupervised representation learning with deep convolutional generative adversarial networks.
\newblock In {\em ICLR}, 2016.

\bibitem{DCGAN}
Alec Radford, Luke Metz, and Soumith Chintala.
\newblock Unsupervised representation learning with deep convolutional generative adversarial networks.
\newblock In {\em ICLR}, 2016.

\bibitem{rezende2015variational}
Danilo Rezende and Shakir Mohamed.
\newblock Variational inference with normalizing flows.
\newblock In {\em ICML}, 2015.

\bibitem{saito2021NIPS}
Kuniaki Saito, Donghyun Kim, and Kate Saenko.
\newblock Openmatch: Open-set semi-supervised learning with open-set consistency regularization.
\newblock In {\em NeurIPS}, 2021.

\bibitem{ScheirerTPAMI2012}
Walter~J Scheirer, Anderson de Rezende~Rocha, Archana Sapkota, and Terrance~E Boult.
\newblock Toward open set recognition.
\newblock {\em IEEE TPAMI}, 35(7):1757--1772, 2012.

\bibitem{ScheirerTPAMI2014}
Walter~J Scheirer, Lalit~P Jain, and Terrance~E Boult.
\newblock Probability models for open set recognition.
\newblock {\em IEEE TPAMI}, 36(11):2317--2324, 2014.

\bibitem{vazeICLR2022}
Sagar Vaze, Kai Han, Andrea Vedaldi, and Andrew Zisserman.
\newblock Open-set recognition: A good closed-set classifier is all you need.
\newblock In {\em ICLR}, 2022.

\bibitem{CUB200}
Catherine Wah, Steve Branson, Peter Welinder, Pietro Perona, and Serge Belongie.
\newblock {The Caltech-UCSD Birds-200-2011 Dataset}.
\newblock Technical report, California Institute of Technology, 2011.

\bibitem{wang2021can}
Haoran Wang, Weitang Liu, Alex Bocchieri, and Yixuan Li.
\newblock Can multi-label classification networks know what they don’t know?
\newblock In {\em NeurIPS}, 2021.

\bibitem{wangICCV2021}
Yezhen Wang, Bo Li, Tong Che, Kaiyang Zhou, Ziwei Liu, and Dongsheng Li.
\newblock Energy-based open-world uncertainty modeling for confidence calibration.
\newblock In {\em ICCV}, 2021.

\bibitem{wang2021energy}
Yezhen Wang, Bo Li, Tong Che, Kaiyang Zhou, Ziwei Liu, and Dongsheng Li.
\newblock Energy-based open-world uncertainty modeling for confidence calibration.
\newblock In {\em ICCV}, 2021.

\bibitem{FGVR_Survey2021}
Xiu-Shen Wei, Yi-Zhe Song, Oisin Mac~Aodha, Jianxin Wu, Yuxin Peng, Jinhui Tang, Jian Yang, and Serge Belongie.
\newblock Fine-grained image analysis with deep learning: A survey.
\newblock {\em IEEE TPAMI}, 2021.
\newblock Early Access.

\bibitem{welling2011bayesian}
Max Welling and Yee~W Teh.
\newblock Bayesian learning via stochastic gradient langevin dynamics.
\newblock In {\em ICML}, 2011.

\bibitem{WongCoRL2020}
Kelvin Wong, Shenlong Wang, Mengye Ren, Ming Liang, and Raquel Urtasun.
\newblock Identifying unknown instances for autonomous driving.
\newblock In {\em CoRL}, 2020.

\bibitem{xia2021adversarial}
Ziheng Xia, Penghui Wang, Ganggang Dong, and Hongwei Liu.
\newblock Adversarial motorial prototype framework for open set recognition.
\newblock {\em arXiv preprint arXiv:2108.04225}, 2021.

\bibitem{xiaoICLR2021}
Zhisheng Xiao, Karsten Kreis, Jan Kautz, and Arash Vahdat.
\newblock Vaebm: A symbiosis between variational autoencoders and energy-based models.
\newblock In {\em ICLR}, 2021.

\bibitem{xu2016webly}
Zhe Xu, Shaoli Huang, Ya Zhang, and Dacheng Tao.
\newblock Webly-supervised fine-grained visual categorization via deep domain adaptation.
\newblock {\em IEEE TPAMI}, 40(5):1100--1113, 2016.

\bibitem{yoshihashiCVPR2019}
Ryota Yoshihashi, Wen Shao, Rei Kawakami, Shaodi You, Makoto Iida, and Takeshi Naemura.
\newblock Classification-reconstruction learning for open-set recognition.
\newblock In {\em CVPR}, 2019.

\bibitem{yu2018hierarchical}
Chaojian Yu, Xinyi Zhao, Qi Zheng, Peng Zhang, and Xinge You.
\newblock Hierarchical bilinear pooling for fine-grained visual recognition.
\newblock In {\em ECCV}, 2018.

\bibitem{yu2022latent}
Peiyu Yu, Sirui Xie, Xiaojian Ma, Baoxiong Jia, Bo Pang, Ruigi Gao, Yixin Zhu, Song-Chun Zhu, and Ying~Nian Wu.
\newblock Latent diffusion energy-based model for interpretable text modeling.
\newblock In {\em ICML}, 2022.

\bibitem{Yue_2021_CVPR}
Zhongqi Yue, Tan Wang, Qianru Sun, Xian-Sheng Hua, and Hanwang Zhang.
\newblock Counterfactual zero-shot and open-set visual recognition.
\newblock In {\em CVPR}, 2021.

\bibitem{zhang2021styleswin}
Bowen Zhang, Shuyang Gu, Bo Zhang, Jianmin Bao, Dong Chen, Fang Wen, Yong Wang, and Baining Guo.
\newblock Styleswin: Transformer-based gan for high-resolution image generation.
\newblock In {\em CVPR}, 2022.

\bibitem{StyleSwin}
Bowen Zhang, Shuyang Gu, Bo Zhang, Jianmin Bao, Dong Chen, Fang Wen, Yong Wang, and Baining Guo.
\newblock Styleswin: Transformer-based gan for high-resolution image generation.
\newblock In {\em CVPR}, 2022.

\bibitem{zhang2021multi}
Fan Zhang, Meng Li, Guisheng Zhai, and Yizhao Liu.
\newblock Multi-branch and multi-scale attention learning for fine-grained visual categorization.
\newblock In {\em ACM MM}, 2021.

\bibitem{zhang2018audio}
Hua Zhang, Xiaochun Cao, and Rui Wang.
\newblock Audio visual attribute discovery for fine-grained object recognition.
\newblock In {\em AAAI}, 2018.

\bibitem{zhang2020hybrid}
Hongjie Zhang, Ang Li, Jie Guo, and Yanwen Guo.
\newblock Hybrid models for open set recognition.
\newblock In {\em ECCV}, 2020.

\bibitem{zhang2016spda}
Han Zhang, Tao Xu, Mohamed Elhoseiny, Xiaolei Huang, Shaoting Zhang, Ahmed Elgammal, and Dimitris Metaxas.
\newblock Spda-cnn: Unifying semantic part detection and abstraction for fine-grained recognition.
\newblock In {\em CVPR}, 2016.

\bibitem{zhangNIPS2021}
Jing Zhang, Jianwen Xie, Nick Barnes, and Ping Li.
\newblock Learning generative vision transformer with energy-based latent space for saliency prediction.
\newblock In {\em NeurIPS}, 2021.

\bibitem{zhang2016picking}
Xiaopeng Zhang, Hongkai Xiong, Wengang Zhou, Weiyao Lin, and Qi Tian.
\newblock Picking deep filter responses for fine-grained image recognition.
\newblock In {\em CVPR}, 2016.

\bibitem{zheng2019learning}
Heliang Zheng, Jianlong Fu, Zheng-Jun Zha, and Jiebo Luo.
\newblock Learning deep bilinear transformation for fine-grained image representation.
\newblock In {\em NeurIPS}, 2019.

\bibitem{zhou2017places}
Bolei Zhou, Agata Lapedriza, Aditya Khosla, Aude Oliva, and Antonio Torralba.
\newblock Places: A 10 million image database for scene recognition.
\newblock {\em IEEE TPAMI}, 40(6):1452--1464, 2017.

\bibitem{zhou2021learning}
Da-Wei Zhou, Han-Jia Ye, and De-Chuan Zhan.
\newblock Learning placeholders for open-set recognition.
\newblock In {\em CVPR}, 2021.

\end{thebibliography}
}

\clearpage
\begin{appendix}

\begin{figure*}[t]
    \centering
    \includegraphics[width=0.95\linewidth]{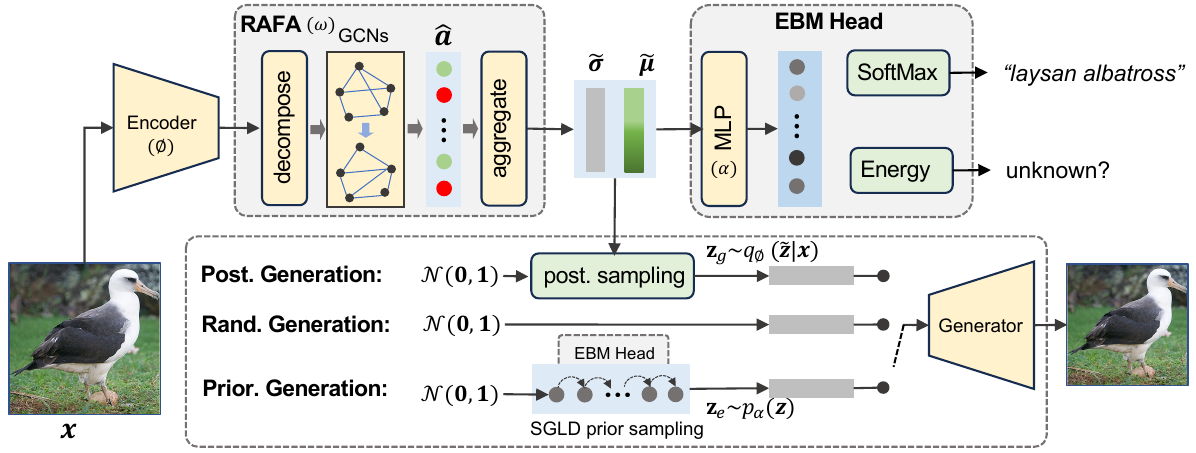}
    \caption{\textbf{Inference Pipeline.} Given a test image, the visual encoder and RAFA module produce the variational posterior distribution of $\mathbf{z}$ which is governed by mean vector $\tilde{\boldsymbol{\mu}}$ and variance vector $\tilde{\boldsymbol{\sigma}}$. The mean vector is used as input of the EBM head for identifying if the image is known or unknown by energy function and softmax classifier for recognition when it is from known classes. For the image generation based on the trained generator, we provide three ways to sample the input latent features: 1) sampling from variational posterior distribution given the image, 2) random sampling from noise $\mathcal{N}(\mathbf{0}, \mathbf{1})$, and 3) sampling from the EBM prior distribution by SGLD sampler.}
    \label{fig:infer}
\end{figure*}

\section*{Appendix}

\appendix


\section{Derivations of Attribute Condition}

To explain why the proposed probabilistic model can be decomposed into an attribute-conditioned SVEBM, here we give the detailed derivation steps as below.
\begin{equation}
    \begin{split}
        p(\mathbf{x},\mathbf{z},\mathbf{a},\mathbf{y}) &= p(\mathbf{y}|\mathbf{a},\mathbf{z})p(\mathbf{a}|\mathbf{z})p(\mathbf{z})p(\mathbf{x}|\mathbf{z}) \\
        &= p(\mathbf{y}|\mathbf{a},\mathbf{x},\mathbf{z}) p(\mathbf{a}|\mathbf{z}) p(\mathbf{x},\mathbf{z}) \\
        &= p(\mathbf{x},\mathbf{y},\mathbf{z}|\mathbf{a}) p(\mathbf{a}|\mathbf{z})
    \end{split}
\end{equation}
Here, the 1st row is the proposed model that shows the relationship between $\mathbf{x},\mathbf{z},\mathbf{y}$ and attribute $\mathbf{a}$. In the 2nd row, we have $p(\mathbf{y}|\mathbf{a},\mathbf{z})=p(\mathbf{y}|\mathbf{a},\mathbf{x},\mathbf{z})$ because $\mathbf{y}$ is conditionally independent of $\mathbf{x}$ given both $\mathbf{a}$ and $\mathbf{z}$, and we also have $p(\mathbf{z})p(\mathbf{x}|\mathbf{z})=p(\mathbf{x},\mathbf{z})$. In the 3rd row, $p(\mathbf{y}|\mathbf{a},\mathbf{x},\mathbf{z})p(\mathbf{x},\mathbf{z}) = p(\mathbf{x},\mathbf{y},\mathbf{z}|\mathbf{a})$ because $\mathbf{y}$ and $(\mathbf{x},\mathbf{z})$ are conditionally independent given the attribute $\mathbf{a}$. 

Eventually, taking the logarithm on both sides, we reach the following distinctive objectives:
\begin{equation}
    \log p(\mathbf{x},\mathbf{z},\mathbf{a},\mathbf{y})=\log p(\mathbf{x},\mathbf{z},\mathbf{y}|\mathbf{a}) + \log p_{\omega}(\mathbf{a}|\mathbf{z}),
\end{equation}
which implies that if $p(\mathbf{x},\mathbf{z},\mathbf{y})$ is assumed to be a SVEBM, it should be conditioned on attribute $\mathbf{a}$ which is determined by latent feature $\mathbf{z}$.

\section{More Implementation Details}

\subsection{Network Architecture} 
\begin{figure*}
    \centering
    \includegraphics[width=\linewidth]{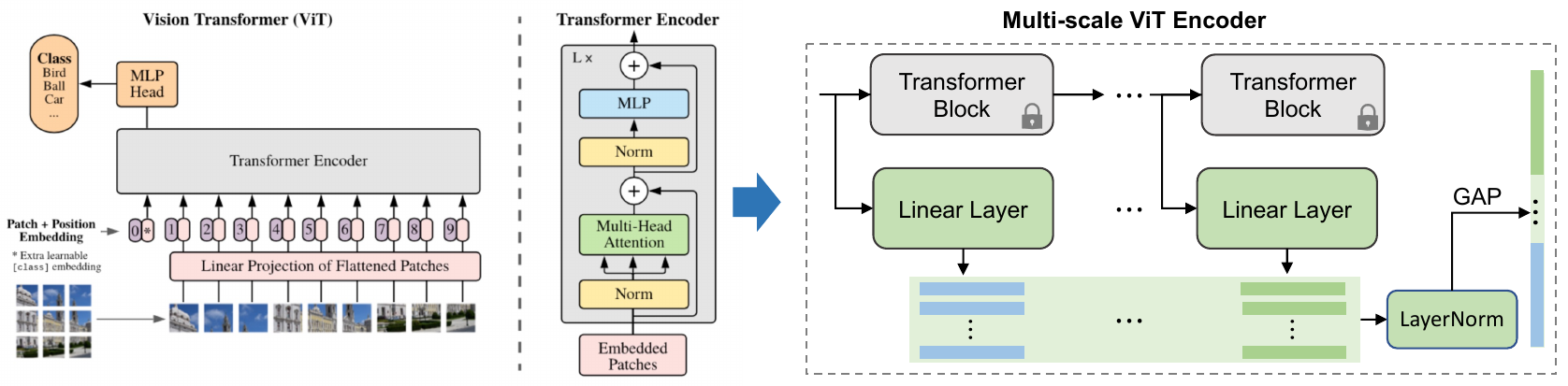}
    \caption{\small \textbf{Multi-scale Vision Transformer}. Based on pre-trained ViT~\cite{vitICLR2021}, we freeze all Transformer Blocks and introduce multiple \emph{Linear Layer} to embed the hierarchical features of ViT into the same embedding space. Followed by a layer normlization (\emph{LayerNorm}) and global average pooling (GAP), the hierarchical ViT features are aggregated as the global representation of the input image.}
    \label{supfig:vit}
\end{figure*}

For our vanilla model, we use pre-trained ResNet-50 as the encoder $q_{\phi1}(\mathbf{z}|\mathbf{x})$. We follow~\cite{vazeICLR2022} to use the pre-trained weights on Places~\cite{zhou2017places} dataset by the unsupervised learning method MoCo-v2~\cite{MoCov2} such that no class supervision will leak into the pre-trained model. For our vision transformer (ViT) based model variants, since ViT is inevitable to be pre-trained, we propose to freeze part of the ViT backbone parameters and learn new learnable branches from scratch or finetune part of layers. Specifically, for the ViT-based encoder, we freeze all the Transformer Blocks in ViT and learn the newly introduced multi-scale side branches (see Fig.~\ref{supfig:vit}). Similar to~\cite{cai2022open}, the ViT backbone we used is the ViT/B-16 ($16\times 16$ patches) with input image size of $384\times 384$. For the vanilla generator DCGAN~\cite{DCGAN}, we pre-train it on the training set for 200 epochs. For the ViT-based generator StyleSwin~\cite{StyleSwin}, we pre-train it on the training set and only fine-tune the first three \emph{EqualLinear} layers and one \emph{StyleBlock} layer. Limited by the GPU resources, we only pre-train the StyleSwin using images with size $256\times 256$ for 500K iterations on CUB and 340K iterations on the DeepFashion dataset, respectively. In Fig.~\ref{fig:infer}, we provide the inference pipeline of our model for fine-grained OSR and image generation.

\subsection{Training Algorithm} 
The training algorithm of the proposed method is summarized in Algorithm~\ref{alg:train}. Consider that in the early training stage, the latent features $\tilde{\mathbf{z}}$ are not sufficiently discriminative. Therefore, we apply the UVOS after $T_{\text{uvos}}$ discriminative training iterations such that the class-wise density estimation is more stable. Similary, we apply the generative training after $T_{\text{gen}}$ discriminative training iterations. In practice, we set both $T_{\text{uvos}}$ and $T_{\text{gen}}$ to to 200 epochs on CUB and TinyImageNet dataset. We note that in SVEBM, the variational inference model $q_{\phi}$, the generative model $p_{\beta}$, and the prior model $p_{\alpha}$ are learned by three gradient descent processes $\text{GD}(\eta)$ where $\eta$ is the initial learning rate. Therefore, we propose to learn the encoder $q_{\phi}(\mathbf{z}|\mathbf{x})$ by $\text{GD}(\eta_0)$, learn the EBM prior model $p_{\alpha}(\tilde{\mathbf{z}},\mathbf{y})$ and UVOS model  $p_{\theta}(\mathbf{s}|\tilde{\mathbf{z}},\mathbf{z}^+)$ by $\text{GD}(\eta_1)$, learn the RAFA model $p_{\omega}(\mathbf{a}|\mathbf{z})$ by $\text{GD}(\eta_2)$, and learn the generator $p_{\beta}(\mathbf{x}|\tilde{\mathbf{z}})$ by $\text{GD}(\eta_3)$, respectively. For each GD process, we use Adam optimizer with weight decay. The learning rates are scheduled by one-time warmup and cosine schedulers with two learning rate restarts. As suggested by~\cite{vazeICLR2022}, we use 10 epochs of warmup training and cosine restarts the learning rate at the 200-th and 400-th epoch within totally 600 training epochs. In the SGLD sampling procedure, we set the number of sampling steps $T_{\text{SGLD}}$ to 100 and step size 0.4.

\begin{algorithm*}[t]
  \caption{Training Algorithm.}
  \label{alg:train}
\begin{algorithmic}
  \STATE {\bfseries Input:} Observed labeled data~$\{(x_i,y_i,a_i)\}_{i=1}^{N}$, learning iterations~$T$, generative learning start $T_{\text{gen}}$, UVOS learning start $T_{\text{uvos}}$, learning rates~$(\eta_0,\eta_1,\eta_2,\eta_3)$, loss weights for RAFA, UVOS, and CMI modules $(\lambda_0,\lambda_1,\lambda_2)$, batch size $B$, number of UVOS samples $H$, initial parameters~$(\alpha_0, \beta_0, \phi_0,\omega_0,\theta_0)$, and number of SGLD steps $T_{\text{SGLD}}$. 
    \STATE {\bfseries Output:} $(\alpha_{T}, \beta_{T}, \phi_{T}, \omega_{T},\theta_{T})$.
  \FOR{$t=0$ {\bfseries to} $T-1$}
  \STATE {\bfseries 1. mini-batch:} Sample a batch of data $\{ x_i, y_i,a_i\}_{i=1}^{B}$.
  \STATE {\bfseries 2. update Gaussian densities:} \\
  \;\;\;\; If $t \leq T_{\text{uvos}}$, update class-wise density statistics $(\hat{\boldsymbol{\mu}}_k,\hat{\mathbf{P}}_k)_t$. \\
  \STATE {\bfseries 3. learn discriminative models ($\phi$, $\alpha$, $\omega$, $\theta$):} \\
  \;\;\;\; Compute $\mathcal{L}(\phi,\alpha,\omega) = -\frac{1}{B}\sum_{i=1}^{B}\left[\log p_{\alpha_t}(\mathbf{y}_i| \tilde{\mathbf{z}}_i= \mu_{\phi_t,\omega_t}(\mathbf{x}_i)) - \lambda_0 \log p_{\omega_t}(\mathbf{a}_i|\mathbf{z}=\mu_{\phi_t}(\mathbf{x}_i))\right]$.  \\
  \;\;\;\; If $t > T_{\text{uvos}}$, sampling $\mathbf{z}^+\in \mathcal{V}^+$ and compute $\mathcal{L}_{\text{uvos}}(\theta)=-\frac{\lambda_1}{B+H}\sum_{i=1}^{B+H}\log p_{\theta}(\mathbf{s}|\tilde{\mathbf{z}}, \mathbf{z}^+)$. \\
  \;\;\;\; Compute joint loss: $\mathcal{L}_d(\alpha,\phi,\omega,\theta) = \mathcal{L}(\alpha,\phi,\omega) + \mathcal{L}_{\text{uvos}}(\theta)$. \\
  \;\;\;\; Update: $\phi_{t+1} = \phi_t - \eta_0 \nabla_\phi \mathcal{L}_d$, \;\; $(\alpha,\theta)_{t+1} = (\alpha,\theta)_t - \eta_1 \nabla_{\alpha,\theta} \mathcal{L}_d$, \;\; $\omega_{t+1} = \omega_t - \eta_2 \nabla_{\omega}\mathcal{L}_d$.
  
  \IF{$t \geq T_{\text{gen}}$}
    \STATE {\bfseries 4. prior sampling:} Randomly sampling a batch of $\mathbf{z}_0\sim \mathcal{N}(\mathbf{0},\mathbf{1})$, and use $T_{\text{SGLD}}$ steps to sample from the EBM prior by SGLD, \ie, $\mathbf{z}^{(e)}\sim p_{\alpha}(\tilde{\mathbf{z}}|\mathbf{z}_0)$ .   
    \STATE {\bfseries 5. posterior sampling:} Use the inference network and reparameterization trick to sample from vatiational posterior, \ie, $\mathbf{z}^{(g)} \sim q_{\phi,\omega}(\tilde{\mathbf{z}}|\mathbf{x})$.
    \STATE {\bfseries 6. learn prior model:} \\
    \;\;\;\; Compute loss $\mathcal{L}(\alpha) = \frac{1}{B}\sum_{i=1}^{B} [E_{\alpha_t}(\mathbf{z}_i^{(g)}) -  E_{\alpha_t}(\mathbf{z}_i^{(e)})] + \lambda_2 \mathcal{CI}_{\alpha_t}(\mathbf{z}^{(g)},\mathbf{y})$. \\
    \;\;\;\; Update $\alpha_{t+1} = \alpha_t - \eta_0\nabla_\alpha \mathcal{L}(\alpha)$. 
    \STATE {\bfseries 7. learn generative models:} \\
    \;\;\;\; Compute loss $\mathcal{L}_g(\phi,\omega,\beta) = -\frac{1}{B}\sum_{i=1}^{B}[\log p_{\beta_t}(\mathbf{x}_i|\mathbf{z}^{(g)}_i) + \mathbb{D}_{\text{KL}}\infdivx{(q_{\phi_t,\omega_t}(\tilde{\mathbf{z}}|\mathbf{x}_i)}{p_0(\mathbf{z})} + E_{\alpha_t}(\mathbf{z}^{(g)}_i)] + \lambda_2 \mathcal{CI}_{\alpha_t}(\mathbf{z}^{(g)},\mathbf{y})$. \\
    \;\;\;\; Update: $\phi_{t+1} = \phi_t - \eta_0 \nabla_\phi \mathcal{L}_g$, \;\; $\omega_{t+1} = \omega_t - \eta_2 \nabla_{\omega}\mathcal{L}_g$, \;\; $\beta_{t+1} = \beta_{t} - \eta_3 \nabla_{\beta} \mathcal{L}_g$.
  \ENDIF
  \STATE {\bfseries 8. update learning rate:} $(\eta_0,\eta_1,\eta_2,\eta_3)\leftarrow \text{CosineWarmupRestart}(\eta_0,\eta_1,\eta_2,\eta_3)$
  \ENDFOR
\end{algorithmic}
\end{algorithm*}

\section{More Qualitative Results}

\begin{figure}[t]
    \centering
    \subcaptionbox{Ours (R50)\label{supfig:tsne_cub_r50dcgan}}{
        \includegraphics[width=0.485\textwidth]{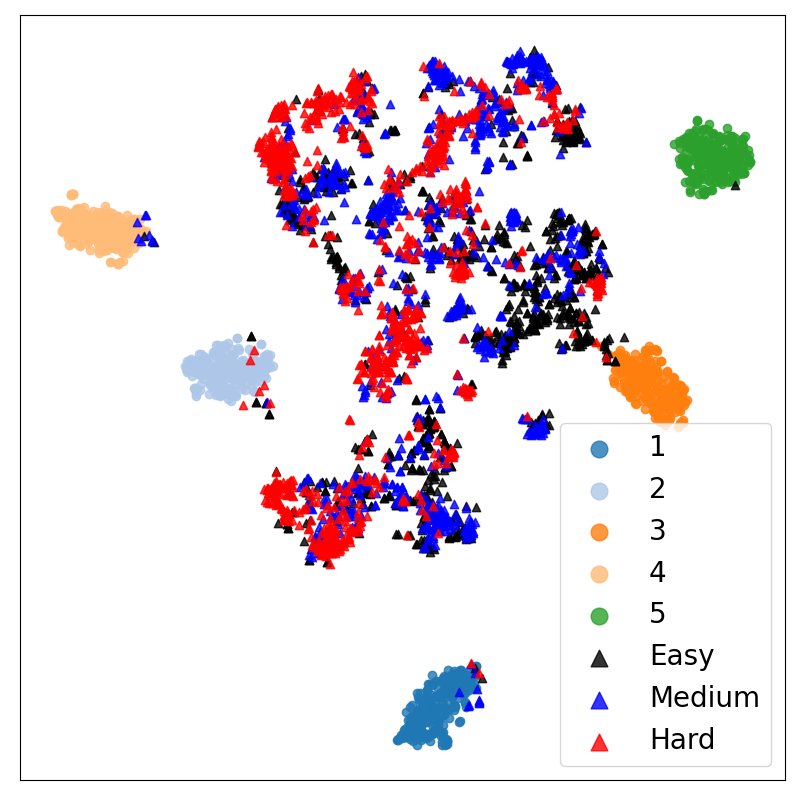}
    }
    \hspace{-3mm}
    \subcaptionbox{Ours (ViT)\label{supfig:tsne_cub_vitdcgan}}{
        \includegraphics[width=0.485\textwidth]{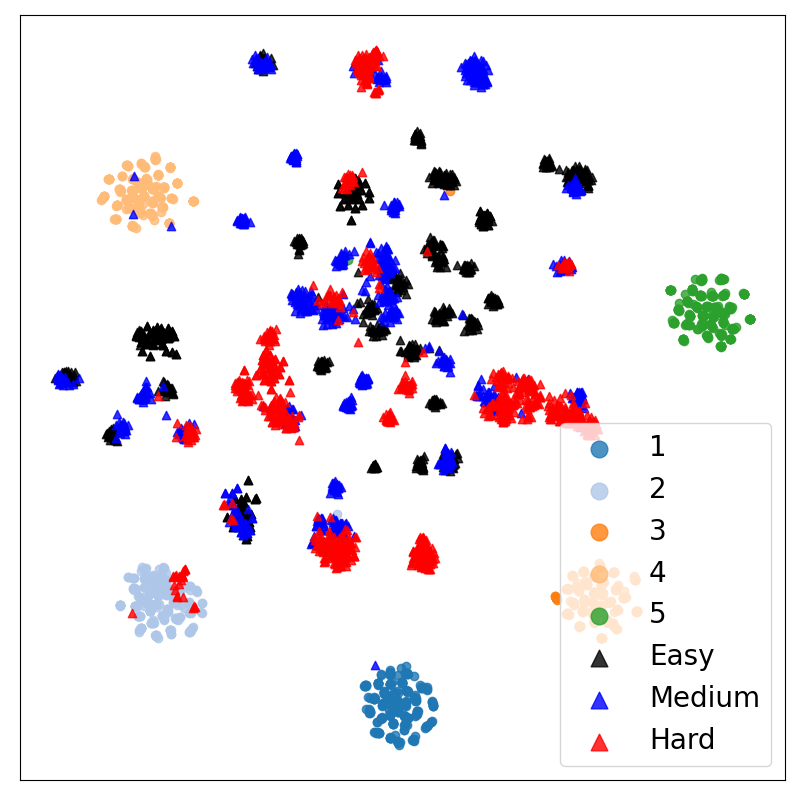}
    }
    \caption{Feature embedding by tSNE on \textbf{CUB} dataset with perplexity of 50. We randomly select 5 known classes with 256 images
for each, and 1024 images for each difficulty level of the unknown.}
    \label{supfig:tsne_cub}
\end{figure}

\begin{figure}[t]
    \centering
    \subcaptionbox{Ours (R50)\label{supfig:tsne_tin_r50dcgan}}{
        \includegraphics[width=0.485\textwidth]{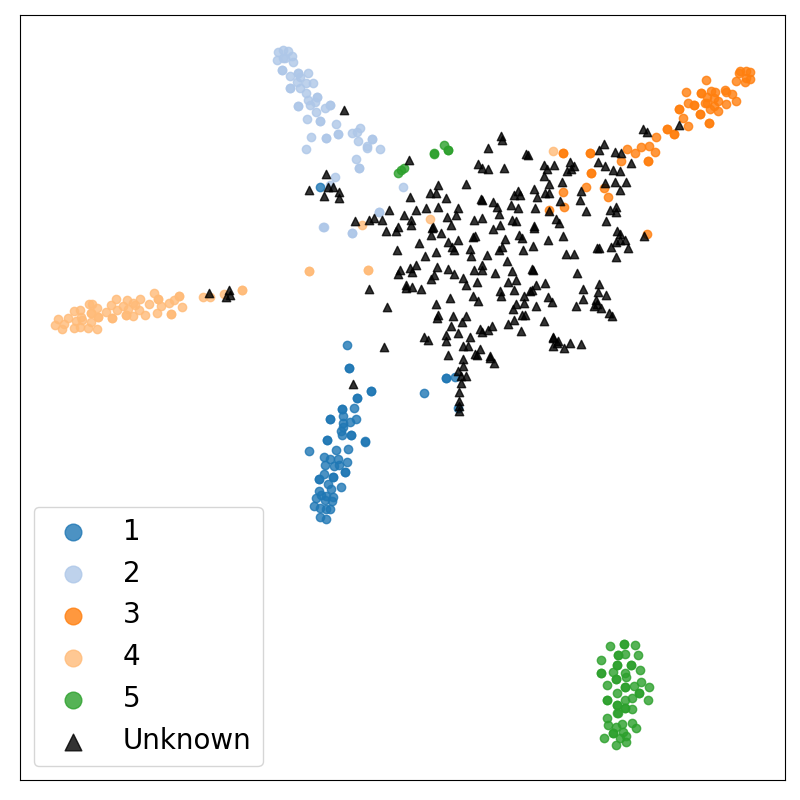}
    }
    \hspace{-3mm}
    \subcaptionbox{Ours (ViT)\label{supfig:tsne_tin_vitdcgan}}{
        \includegraphics[width=0.485\textwidth]{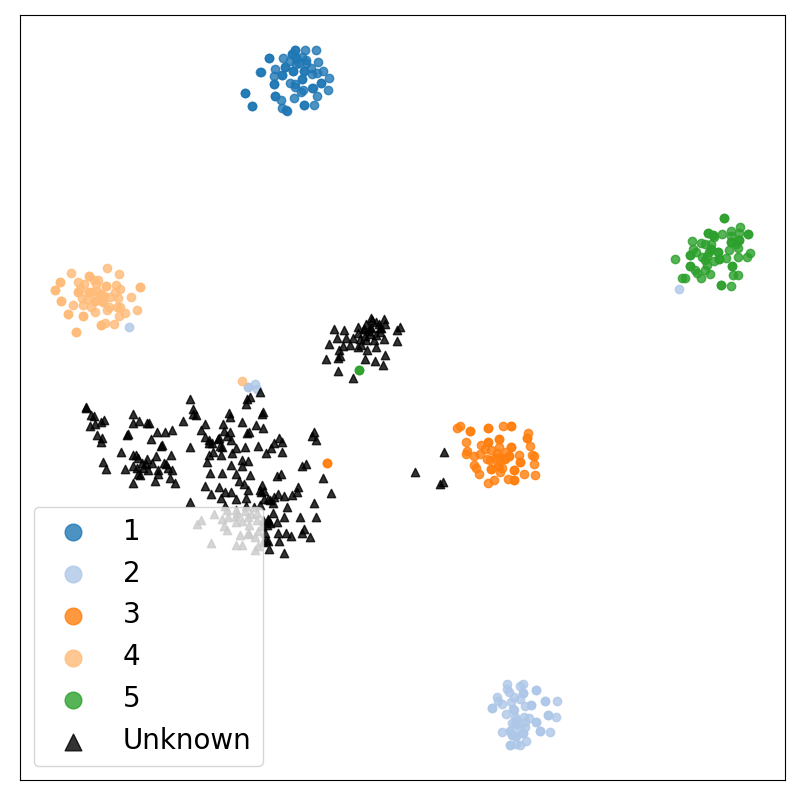}
    }
    \caption{Feature embedding by tSNE on \textbf{TinyImageNet} dataset  with perplexity of 30. We randomly select 5 known classes with 64 images
for each, and 256 images for the unknown.}
    \label{supfig:tsne_tin}
\end{figure}

\subsection{Feature Embedding} In Fig.~\ref{supfig:tsne_cub} and~\ref{supfig:tsne_tin}, we show the learned feature embeddings of the proposed two model variants by tSNE on CUB and TinyImageNet datasets. In each figure, we randomly select five classes with 256 images for each from closed-set training data as the known data, and we randomly select 1024 images for each difficulty level of the unknown classes. As expected, the model variants with ViT encoder could achieve much better separation between the known classes and different difficulty levels of the unknown. Moreover, for different difficulty levels of the unknown, the ViT-based embeddings could form a clustering effect for those unknown when compared with the R50-based embeddings. This observation indicates the potential of vision transformers for discovering unknown classes in other related research topics, such as novel class discovery, life-long learning, and open-world learning. We also notice that for some known classes, there are several data points of the unknown that appear at the edge of the class cluster. This could be the side effect of the UVOS that synthesizes virtual unknowns by sampling at the low-density area of known class data. 


\subsection{Image Generation}

\begin{figure*}[t]
    \centering
    \subcaptionbox{Real images: $\mathbf{x}\sim p_{\text{data}}(\mathbf{x})$\label{supfig:fashion_real}}{
        \includegraphics[width=0.48\linewidth]{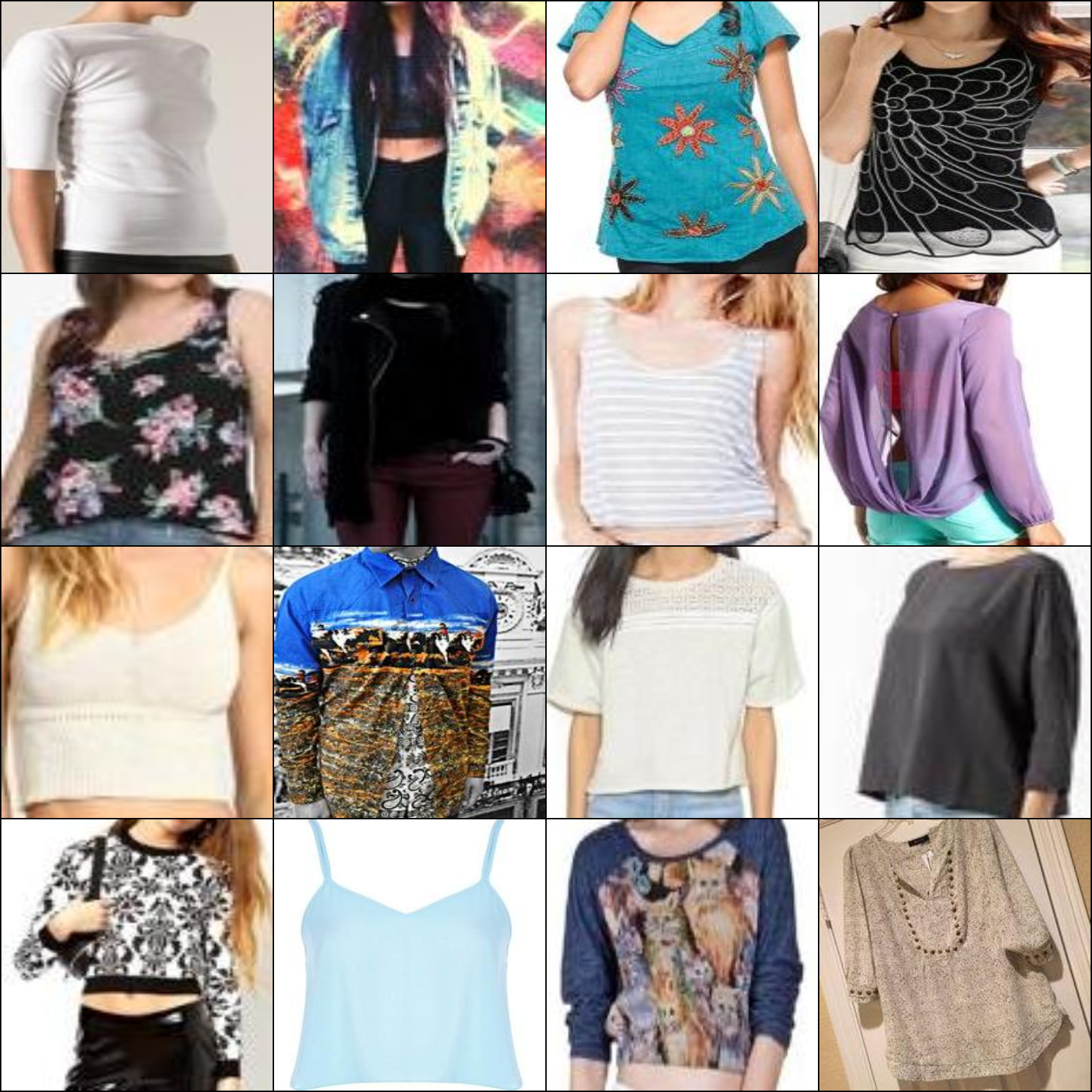}
    }
    \hspace{-3mm}
    \subcaptionbox{Fake images (random): $\mathbf{x}\!=\!f_{\beta^*}(\mathbf{z}),\; \mathbf{z}\!\sim\! \mathcal{N}(0,1)$\label{supfig:fashion_pretrain}}{
        \includegraphics[width=0.48\linewidth]{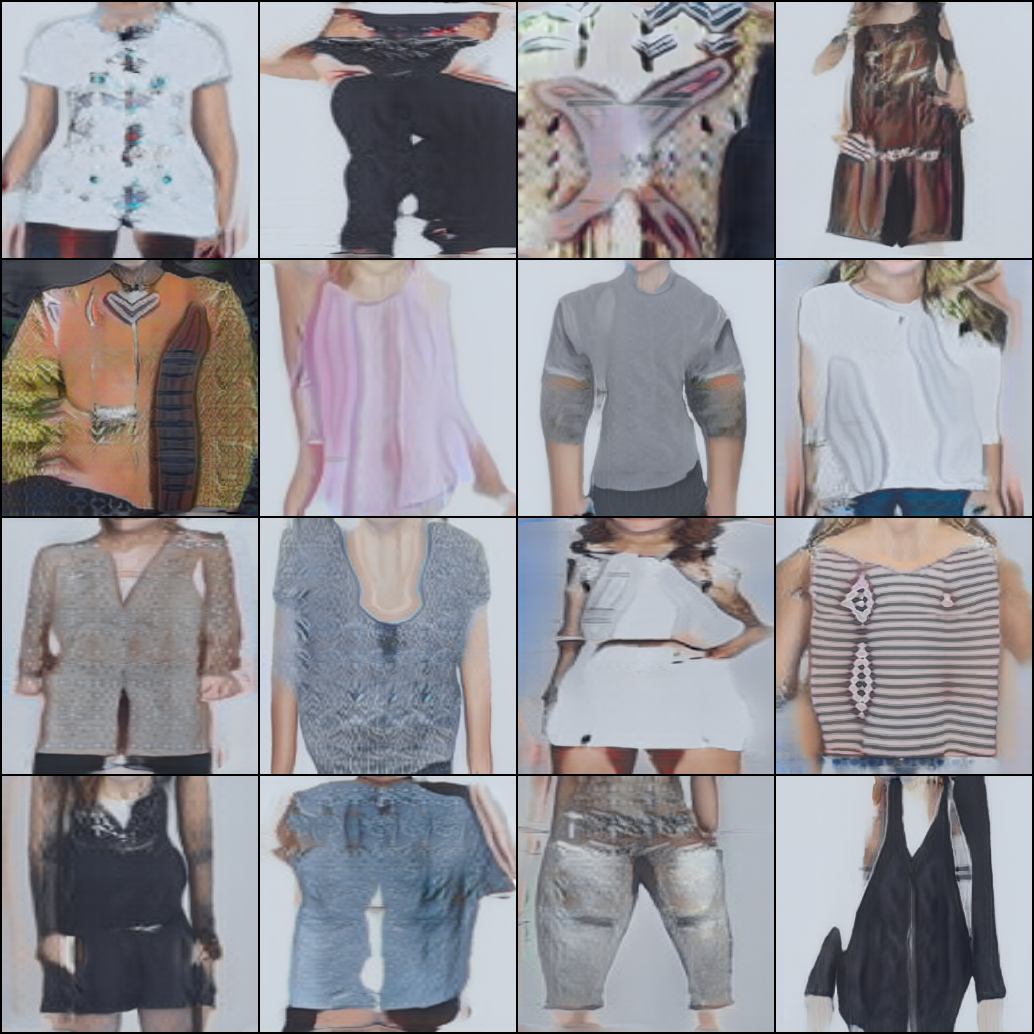}
    }
    \hspace{-3mm}
    \subcaptionbox{Fake images (posterior): $\mathbf{x}\!=\!f_{\beta}(\mathbf{z}),\;\mathbf{z}\!\sim\!q_{\phi}(\mathbf{z}|\mathbf{x})$ \label{supfig:fashion_post}}{
        \includegraphics[width=0.48\linewidth]{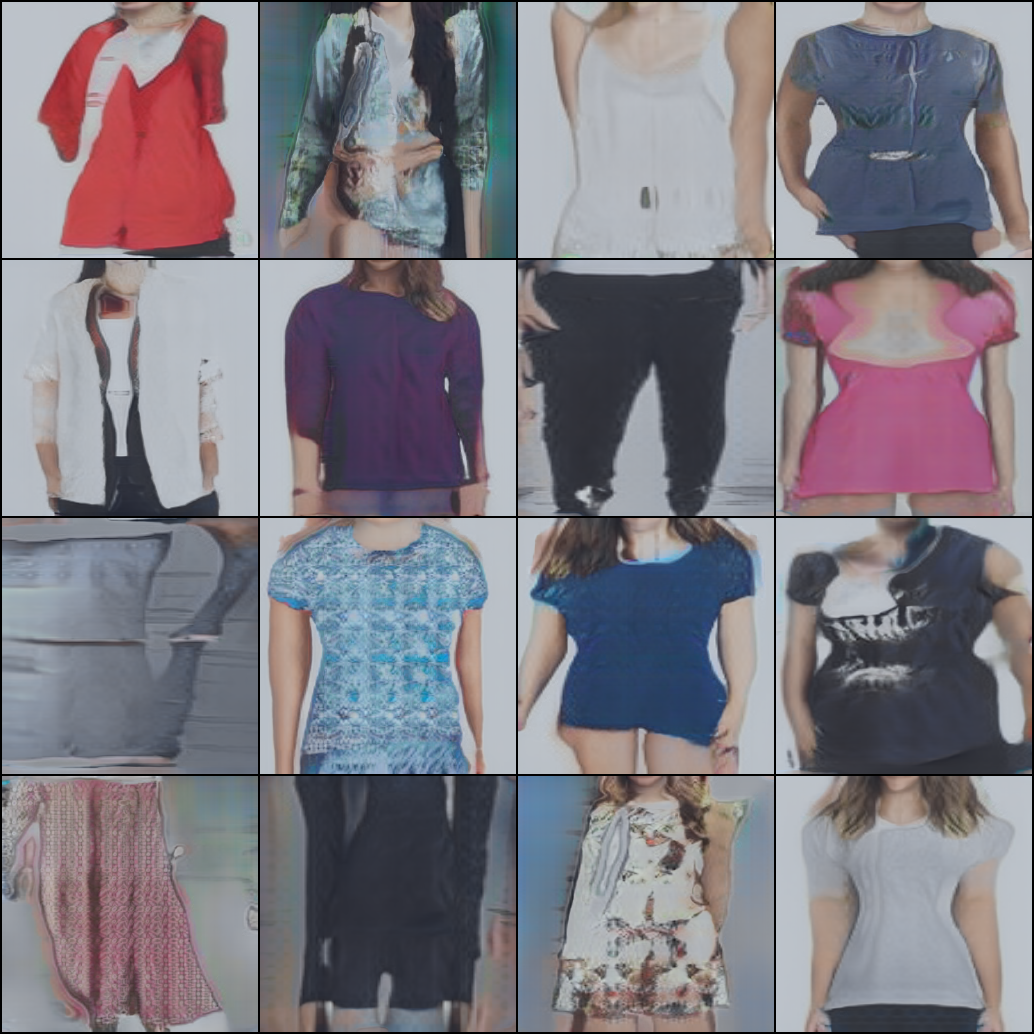}
    }
    \hspace{-3mm}
    \subcaptionbox{Fake images (prior): $\mathbf{x}\!=\!f_{\beta}(\mathbf{z}),\;\mathbf{z}\!\sim\!p_{\alpha}(\mathbf{\mathbf{z}})$\label{supfig:fashion_prior}}{
        \includegraphics[width=0.48\linewidth]{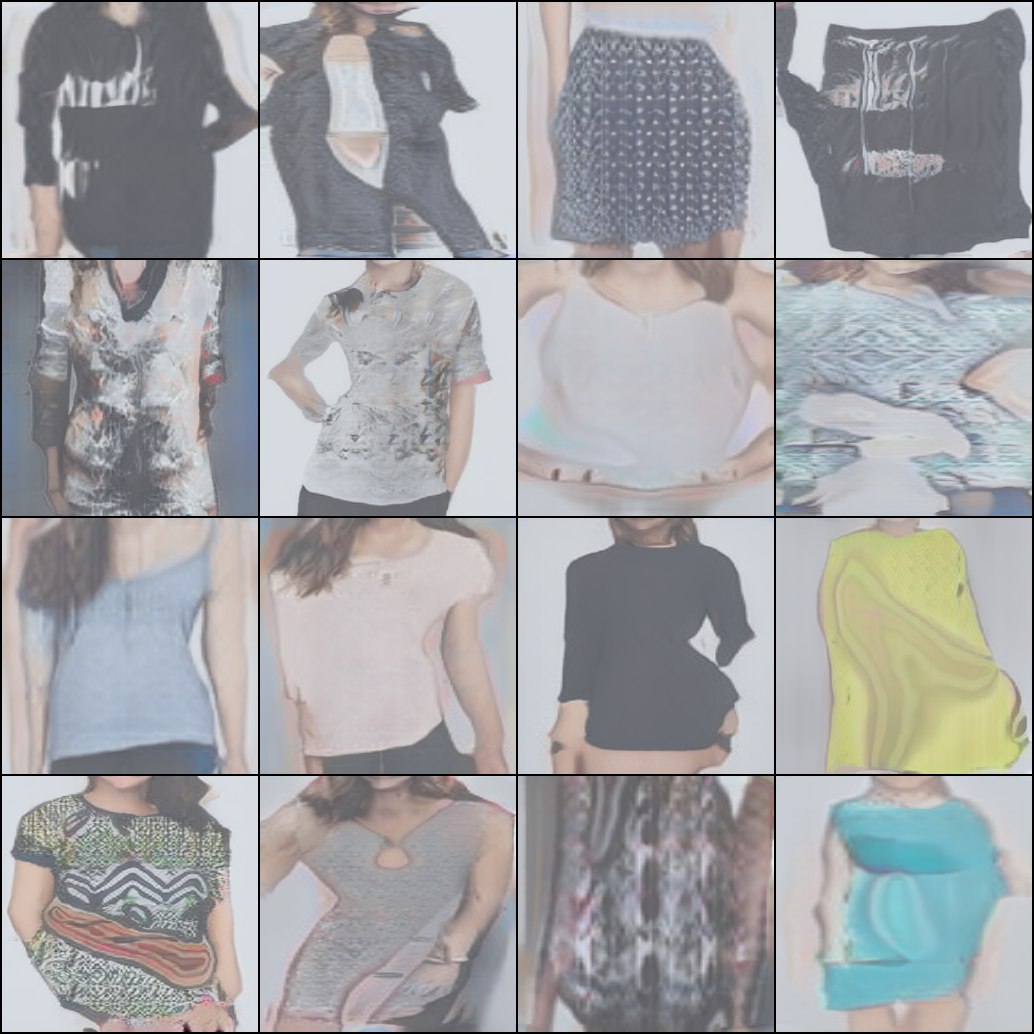}
    }
\captionsetup{font=small,aboveskip=5pt}
  \caption{\textbf{Image Generation on DeepFashion dataset}. All images are with the size $256\times 256$. We compare different methods including the results using only the pre-trained generator (Fig.~\ref{supfig:fashion_pretrain}), results using the features sampled from the variational posterior (Fig.~\ref{supfig:fashion_post}), and the results using features sampled from EBM prior (Fig.~\ref{supfig:fashion_prior}).}
  \label{supfig:gen_fashion}
\end{figure*}

In Fig.~\ref{supfig:gen_fashion}, we show the generated images by different latent sampling techniques on the DeepFashion dataset. 

\end{appendix}

\end{document}